# Beyond Acuity Accuracy: A Multidimensional Framework for Evaluating Patient-Facing Emergency Triage AI Tools

Ruio Heerschap[1,2], Henry Foote[3], Mahmoud Alwakeel[1,4], Zhouhan Qian[1], Jiaqi Wang[1], Elisabeth Bond[1], Yang Yang[1], Jonathan Chong Kai Liew[5], Matthew A. Bergens[6], Monique Arnold[5,7,8], Haoyuan Wang[1], Rui Yang[9], Kunyu Yu[9], Nikita R. Paripati[5,7,8], Diana Pimenta[7], Kyle Naimoli[7], An-Kwok Ian Wong[10,11,12], Nicoleta Economou-Zavlanos[1], João Ricardo Vissoci[6], Nan Liu[1,9,13], Anand Chowdhury[4], Chuan Hong[1,*]

*1 Department of Biostatistics and Bioinformatics, Duke University School of Medicine, Durham, NC, USA*
*2 Leiden University Medical Centre, Leiden, The Netherlands*
*3 Department of Pediatrics, Duke University School of Medicine, Durham, NC, USA*
*4 Department of Medicine, Duke University School of Medicine, Durham, NC, USA*
*5 Perelman School of Medicine, University of Pennsylvania, Philadelphia, PA, USA*
*6 Department of Emergency Medicine, Duke University, Durham, NC, USA*
*7 Hospital of the University of Pennsylvania, Philadelphia, PA, USA*
*8 Children's Hospital of Philadelphia (CHOP), Philadelphia, PA, USA*
*9 Centre for Biomedical Data Science, Duke-NUS Medical School, Singapore, Singapore*
*10 Division of Pulmonary, Allergy, and Critical Care Medicine, Duke University, Durham, NC, USA*
*11 Division of Translational Biomedical Informatics, Duke University, Durham, NC, USA*
*12 Staff Physician, Durham VA Medical Center, NC, USA*
*13 NUS Artificial Intelligence Institute, National University of Singapore, Singapore, Singapore*
** Correspondence to: Chuan Hong, chuan.hong@duke.edu*

## ABSTRACT

Patient-facing AI triage tools may support care-seeking and emergency department prioritization, but require evaluation beyond acuity accuracy alone. We developed SAFE-Triage, a framework integrating human review, automated metrics, LLM-as-judge evaluation, and simulation testing across clinical performance, patient interaction, and clinician-facing summary quality. We applied SAFE-Triage to 40 MIMIC-IV-derived emergency vignettes and a 10,030-case acuity analysis comparing prompted LLM triage (PLT) with rule-augmented prompted LLM triage (rPLT). PLT achieved higher emergency severity index (ESI) accuracy (57.5% vs 42.8%, $P < 0.001$), lower undertriage (13.1% vs 29.8%, $P < 0.001$), and higher clinician ratings across all domains. rPLT produced shorter and more stable outputs with higher automated readability scores. SAFE-Triage identified trade-offs between acuity accuracy, undertriage risk, communication quality, readability, and summary quality, supporting pre-deployment refinement and ongoing monitoring by showing where triage tools are efficient, need improvement, or need clinician oversight.

## INTRODUCTION

Emergency department (ED) crowding has become a critical and growing challenge across healthcare systems internationally. Although ED utilization is not captured in a single standardized global registry, available national and international data indicate increasing demand for emergency care across several healthcare settings.[1–3] For example, U.S. ED visits increased from approximately 145.6 million in 2016 to 155.4 million in 2022.[4] These conditions are consistently associated with adverse outcomes, including increased mortality and reduced quality of care, ultimately leading to suboptimal patient management.[5–7] As demand for emergency services continues to increase, ED crowding places substantial strain on healthcare providers and infrastructure, highlighting the urgent need for strategies that improve patient prioritization and workflow within the emergency setting.[6–9]

Multiple factors contribute to increasing pressure on EDs. Demographic changes, including population growth and ageing, have increased healthcare utilization and consequently ED demand.[7,10] In addition, inappropriate ED use has become more common, as limited access to primary care and increasing health-related anxiety, often amplified by online health information seeking such as cyberchondria, can drive patients to seek emergency care for nonurgent conditions and are associated with higher healthcare utilization.[11,12] Operational limitations further exacerbate overcrowding, as manual triage is time-consuming, varies across providers, and is difficult to scale during periods of high patient influx; combined with persistent staff shortages, these constraints reduce patient throughput and delay timely assessment, with triage times worsening further under crowded conditions.[7,10,13]

Artificial intelligence (AI)-driven triage tools have been proposed to support more appropriate care-seeking, improve patient prioritization, and reduce avoidable ED use. Earlier automated triage approaches often relied on rule-based decision systems, which offered consistency but were limited in their ability to adapt to varied patient presentations, engage in natural conversation, or reason beyond predefined criteria. The emergence of large language models (LLMs) has introduced new possibilities for more flexible and conversational triage approaches.[14,15,16]

Existing applications include both patient-facing pre-ED symptom checkers and in-ED decision-support tools for triage and clinical prioritization.[17] Recent work has also demonstrated the feasibility of LLM-based triage workflows in clinical decision support.[18,19] Patient-facing symptom checkers may additionally improve users' confidence in seeking appropriate care and reduce health-related anxiety, suggesting potential benefits beyond operational efficiency alone.[20,21] In addition to general-purpose LLM-based approaches, dedicated patient-facing symptom checkers and triage tools have also been developed. Prior evaluation has shown encouraging but imperfect triage performance of these systems, with acuity advice matching the gold standard in 63% of cases and agreement with clinicians' decisions in 62% of ED patients.[22,23]

Despite their promise, AI-assisted triage tools raise important safety and implementation concerns, as undertriage may delay treatment for critically ill patients, while overtriage may divert limited emergency resources and worsen ED congestion.[24] Rigorous evaluation is therefore essential to determine whether these tools can provide safe, reliable, and operationally appropriate recommendations across diverse clinical scenarios.[25] Existing evaluations have largely relied on human expert assessment, which is limited by subjectivity and substantial time and resource demands, making results difficult to scale and compare across studies, particularly when evaluation domains, scoring approaches, and reference standards differ.[26] In addition, most evaluations focus primarily on clinical performance, such as diagnostic accuracy and triage accuracy, while giving less attention to patient-facing aspects of the interaction. For tools that communicate directly with patients, safety and usefulness also depend on whether the interaction is clear, coherent, understandable, and sufficient to gather the information needed for safe triage decisions. Evaluation should therefore assess patient-facing communication alongside clinical outcomes. Beyond individual tools, the absence of a unified standard for ED triage AI limits comparability and makes it difficult to determine whether new tools meaningfully improve clinical safety, patient-facing communication, or downstream clinical use.

To address these limitations, we propose **SAFE-Triage** (**S**tandardized **A**ssessment **F**ramework for **E**mergency **Triage tools**), an evaluation framework for triage AI tools integrating human evaluation, auto evaluation, LLM-as-judge evaluation, and simulation evaluation. SAFE-Triage adapts the SCRIBE framework, which combines complementary evaluation methods for the multidimensional assessment of clinical AI outputs, to the specific requirements of emergency triage.[27,28] SAFE-Triage enables assessment beyond acuity accuracy alone, including clinical safety, patient-facing communication, and clinician-facing summary quality. This design also builds on recent studies showing that LLM-judges can achieve promising agreement with human raters when judging AI-generated medical outputs using structured rubrics.[29,30]

In this study, we illustrate SAFE-Triage by evaluating two patient-facing triage tools: prompted LLM triage (PLT) and rule-augmented prompted LLM triage (rPLT). This comparison demonstrates how the framework can provide a coherent, multidimensional, and clinically meaningful evaluation across modules.

# RESULTS

SAFE-Triage was applied to compare PLT and rPLT across 40 MIMIC-IV-derived clinical vignettes, a separate 10,030-case triage classification analysis, and simulations involving rare diseases and must-not-miss conditions (Figure 1 and Table 1).[31] Both tools used the same underlying case information and generated the same output types: diagnosis and differential diagnosis, a patient-facing care recommendation (Table 2), and a clinician-facing SBAR summary with a predicted Emergency Severity Index (ESI) level.

## Clinical vignette characteristics

The 40 MIMIC-IV-derived clinical vignettes served as the shared case set used as input for both triage tools across human evaluation, scorecard-aligned automated metrics, LLM-as-judge evaluation, and repeated-output stability testing. Vignette characteristics are summarized in Table 3. Median patient age was 55.5 years (IQR 33.5–68.0; range 20–85), and the vignette set was evenly distributed by sex, with 20 female and 20 male patients. The cases followed a predefined chief complaint distribution, with abdominal pain, chest pain, dyspnea, fever, headache, back pain, altered mental status, dizziness/lightheadedness, and vomiting/nausea represented in descending order of frequency. All ESI levels were represented, with ESI 1, 2, 3, 4, and 5 accounting for 12.5%, 30.0%, 37.5%, 15.0%, and 5.0% of cases, respectively. The history of present illness sections had a median length of 332 words per case (IQR 244.5–371.3; range 108–585).

## Human evaluation

Human evaluation used a structured six-point scorecard covering clinical performance, patient interaction and dialogue quality, and clinician-facing summary quality (Table 1). Human raters independently rated each triage tool output for each case. Agreement was assessed before score aggregation using exact agreement and within-one-point agreement, defined as ratings that differed by no more than 1 point on the six-point scale. Final human scores were calculated as the mean of the two raters, except for ratings with substantial disagreement, defined as a score difference of 3 or more, for which an adjudicated consensus score was used. Clinician scores served as the primary human reference, while non-clinician ratings were restricted to the patient interaction and dialogue quality domain and were analyzed separately.

Two critical care clinicians (H.F. and M.Al.) independently rated all 40 cases for both tools. Before adjudication, exact clinician inter-rater agreement was 67.9% for PLT and 48.7% for rPLT, and within-one-point agreement was 98.5% and 88.8%, respectively. Substantial disagreement (≥3 points) occurred in 21 of 1,040 ratings (2.0%) and was resolved through adjudication to generate the final clinician reference scores. Detailed agreement scores can be found in Supplementary Information A.

Clinicians rated PLT higher than rPLT across all evaluated dimensions on the 6-point Likert scale (Table 4, Figure 2a-d). For PLT, mean scores were highest for clarity (5.98), user trust and SBAR adherence (both 5.94), and completeness (5.93), and lowest for question sufficiency (5.18), diagnosis accuracy (5.29), and triage recommendation appropriateness (5.35). For rPLT, mean scores were highest for brevity (5.04), medical jargon and SBAR adherence (both 5.00), and clarity (4.96), and lowest for question sufficiency (4.08), triage recommendation appropriateness (4.25), and relevance (4.28). In a supplementary review by two emergency physicians with access to the original scores, PLT was again scored higher than rPLT on all 13 dimensions (mean across dimensions, 5.41 vs. 4.53, compared with 5.75 vs. 4.63 for the critical care clinicians) and in 38 of 40 cases ($p < 0.001$). Emergency physician scores were lower than the original scores mainly for prudence and user trust with both tools. The two emergency physicians' ratings were within one point of each other for 90.2% of PLT ratings and 73.8% of rPLT ratings (Supplementary Information H).

Two patterns were notable. First, both tools received their lowest scores in question sufficiency, diagnosis accuracy, and triage recommendation appropriateness, indicating persistent limitations in the dimensions most directly tied to clinical performance. Second, rPLT retained its highest relative scores in brevity, medical jargon, and SBAR adherence, suggesting concise and standardized outputs, but this did not translate into higher overall clinician-rated performance.

To capture a non-medical, patient-facing perspective, two non-clinician raters (E.B. and J.C.K.L.) assessed only the patient interaction and dialogue quality domain using the same scorecard approach. Exact inter-rater agreement was 44.6% for PLT and 45.8% for rPLT, and within-one-point agreement was 77.9% and 83.3%, respectively. Substantial disagreement occurred in 24 of 480 ratings (5.0%) and was adjudicated to generate the final non-clinician reference scores. Detailed agreement scores can be found in Supplementary Information A.

Across all six patient interaction dimensions, non-clinician raters rated PLT higher than rPLT (Table 4, Figure 2b & 2d). For PLT, mean scores were highest for relevance (5.72), clarity (5.70), and medical jargon (5.58), and lowest for fluency and user trust (both 5.24). For rPLT, mean scores were highest for relevance (5.35), clarity (5.30), and fluency (5.15), and lowest for user trust (3.67), medical jargon (4.30), and coherence (4.95).

Agreement between clinician and non-clinician ratings was assessed for the patient interaction and dialogue dimensions scored by both groups, and was low overall, with Spearman correlations of 0.133 for PLT ($p = 0.039$) and −0.121 for rPLT ($p = 0.061$). Non-clinicians scored PLT lower than clinicians across all six dimensions, while rPLT showed mixed disagreement:

non-clinicians scored user trust and medical jargon lower, whereas clinicians scored relevance, coherence, fluency, and clarity lower. The largest differences were rPLT relevance (non-clinician > clinician by 1.08), rPLT user trust (clinician > non-clinician by 0.94), rPLT medical jargon (clinician > non-clinician by 0.70), and PLT user trust (clinician > non-clinician by 0.70), indicating the greatest divergence around perceived trust and medical language use.

## Auto evaluation

Auto evaluation provided reproducible quantitative assessment alongside human review. It included a separate 10,030-case MIMIC-IV acuity analysis of ESI classification performance, repeated-output stability testing across the 40 clinical vignettes, and scorecard-aligned automated metrics for dimensions that could be assessed quantitatively. Scorecard-aligned metrics were evaluated in reference-free mode for both tools and in reference-based mode only for rPLT, because clinician-generated references were created only for rPLT. Associations between automated metrics and clinician ratings were assessed using Spearman correlations.

In the separate 10,030-case MIMIC-IV triage classification analysis, ESI levels predicted by each triage tool were compared with reference ESI levels using exact accuracy, acceptable accuracy, and directional error rates (Figure 3). Exact accuracy required an exact ESI match, acceptable accuracy allowed an exact match or a prediction one level more conservative, meaning assignment to a triage category one level more urgent than the reference, and directional errors were reported as undertriage or overtriage. Baseline characteristics of the 10,030-case set can be found in Supplementary Information A.

PLT achieved higher exact ESI accuracy than rPLT (57.47% vs. 42.84%; paired difference, 14.63 percentage points; 95% CI, 13.31 to 15.94; $p < 0.001$). PLT also achieved higher acceptable accuracy (86.12% vs. 67.20%; paired difference, 18.92 percentage points; 95% CI, 17.94 to 19.89; $p < 0.001$). The increase from exact to acceptable accuracy for both tools indicates that misclassifications were concentrated between adjacent ESI levels, particularly ESI 2 and ESI 3, whereas larger deviations from the reference label were uncommon.

When errors were partitioned by direction, undertriage occurred less frequently with PLT than with rPLT (13.07% vs. 29.77%; paired difference, −16.70 percentage points; 95% CI, −17.61 to −15.77; $p < 0.001$). Overtriage was slightly more frequent with PLT (29.46% vs. 27.39%; paired difference, 2.07 percentage points; 95% CI, 1.07 to 3.09; $p < 0.001$). In exploratory analyses by presenting complaint, PLT performance for trauma-related complaints (n = 627) was similar to its overall performance (exact accuracy, 58.9%; undertriage, 14.2%), whereas undertriage for acute neurological complaints (n = 551) was higher with both tools (26.3% for PLT and 58.3% for rPLT). rPLT undertriage was also higher for trauma-related complaints (48.6%) (Supplementary Information F). Undertriage was concentrated among encounters with a reference ESI level of 1, of which 67.1% were assigned a less urgent level by PLT and 85.1% by rPLT. PLT undertriage in these encounters was almost entirely by one level (1.9% by two or more levels), whereas rPLT assigned ESI level 3 or lower to 45.1% of them (Supplementary Information I).

We next assessed repeated-output stability by running each of the 40 vignettes 20 times for both tools under identical parameters. Categorical stability for ESI prediction and patient-facing triage recommendation was summarized using modal-category consistency, defined as the proportion of repeated runs producing the most frequent category for that case, while diagnostic stability was summarized using cosine similarity. rPLT showed complete consistency in ESI prediction compared with PLT (mean modal-category consistency, 1.00 vs 0.85; $p < 0.001$). Patient-facing triage recommendations were also highly stable for both tools, with similar modal-category consistency for rPLT and PLT (0.96 vs 0.95; $p = 0.815$). Diagnostic outputs showed high semantic consistency across repeated runs. PLT had slightly higher primary diagnosis stability than rPLT (mean cosine similarity, 0.96 vs 0.94; $p < 0.001$), whereas rPLT had slightly higher stability when primary and differential diagnoses were combined (0.95 vs 0.94; $p = 0.002$). Overall, rPLT produced more stable categorical outputs, particularly for ESI prediction, while both tools showed high stability in patient-facing recommendations and diagnostic text similarity.

For the scorecard-aligned automated evaluation, reference-free metrics were calculated directly from tool outputs without clinician-generated references. Reference-based metrics compared triage tool outputs with clinician-generated references. Because these references were created only for rPLT, reference-based evaluation was performed only for rPLT. Reference-free evaluation showed that PLT produced longer outputs (230.45 vs 124.75 words, $p < 0.001$), lower readability scores (Flesch Reading Ease 72.88 vs 86.99, $p < 0.001$), higher perplexity (16.90 vs 11.79, $p < 0.001$), and lower medical term density (0.2546 vs 0.3064, $p < 0.001$) than rPLT. By the convention used here, rPLT was therefore more fluent, readable, and concise, while medical term density was treated descriptively. Clinician ratings showed the opposite direction for the corresponding dimensions, with PLT rated higher than rPLT for clarity (5.98 vs. 4.96) and fluency (5.88 vs. 4.60).

In the reference-based analysis, rPLT showed strong similarity to clinician-generated references across all metrics, with the highest scores for BERTScore precision (0.975), F1 (0.966), and recall (0.957), and lower scores for BLEU (0.716), METEOR (0.740), and ROUGE-L F1 (0.815).

To assess whether automated metrics reflected human evaluation, scorecard-aligned automated scores were correlated with clinician ratings for the corresponding dimensions using Spearman's ρ. The strongest correlations were observed for relevance via BERTScore F1 ($\rho = 0.322$) and Summary Completeness using ROUGE-Lsum F1 ($\rho = 0.177$). In contrast, clarity via Flesch Reading Ease ($\rho = -0.006$) and fluency via inverse perplexity ($\rho = -0.081$) showed weak or inconsistent associations (Figure 4a).

Automated metric alignment with non-clinician ratings was similarly limited, with mean absolute Spearman correlations of 0.175 for PLT and 0.138 for rPLT across mapped dimensions. The strongest automated-metric correlations with non-clinician ratings were observed for PLT Fluency using inverse perplexity ($\rho = -0.347$) and rPLT Medical Jargon using medical-term density ($\rho = -0.219$). The weakest alignment was observed for rPLT Clarity using Flesch Reading Ease ($\rho = -0.060$) and PLT Medical Jargon using medical-term density ($\rho = -0.085$) (Figure 4b).

## LLM-as-judge evaluation

GPT-5.2 was used as an LLM-judge to score triage outputs across the same dimensions and on the same 1 to 6 scale used by human raters.[32] In reference-free mode, the LLM-judge scored PLT and rPLT using the original case materials and triage tool outputs provided to human raters. In reference-based mode, the LLM-judge additionally received clinician-generated rPLT reference outputs as scoring references and was therefore applied only to rPLT. To account for stochastic variation in LLM-generated ratings, each case and scorecard dimension was independently scored across 7 repeated judging runs, and the median score was used as the final LLM-judge rating.

In reference-free mode, the LLM-judge rated triage outputs more strictly than clinicians overall (Table 4, Figure 2). For PLT, the mean LLM-judge score was 4.54 compared with 5.75 for clinician ratings; for rPLT, the mean LLM-judge score was 4.24 compared with 4.63 for clinician ratings. The largest gaps between LLM-judge and clinician scores for rPLT were observed in completeness (−2.28), question sufficiency (−1.63), and diagnosis accuracy (−1.56). Consistent with clinician evaluation, LLM-judge scores in reference-free mode were higher for PLT than for rPLT.

When clinician-generated reference outputs were added to the LLM-judge prompt for rPLT, the mean LLM-judge score was 4.48, slightly higher than the reference-free mean of 4.24 but still below the clinician mean of 4.63. Reference-based LLM judging was not performed for PLT, as clinician-generated reference outputs were created only for rPLT.

Alignment between LLM-judge and clinician ratings was summarized using mean absolute Spearman correlation across scorecard dimensions. In reference-free mode, overall agreement with clinician ratings was 0.190 for PLT and 0.186 for rPLT; under reference-based judging, agreement for rPLT increased to 0.223. For rPLT under reference-free judging, the strongest dimension-level correlations were diagnosis accuracy ($\rho = 0.479$), question sufficiency ($\rho = 0.440$), and triage recommendation appropriateness ($\rho = 0.312$); the weakest were fluency ($\rho = 0.010$), relevance ($\rho = 0.030$), and brevity ($\rho = 0.043$). Under reference-based judging, the strongest were triage recommendation appropriateness ($\rho = 0.730$), question sufficiency ($\rho = 0.625$), and diagnosis accuracy ($\rho = 0.39$); the weakest were brevity ($\rho = 0.042$), fluency ($\rho = 0.052$), and SBAR adherence ($\rho = 0.057$) (Figure 4a). Adding clinician-generated reference outputs therefore improved LLM-judge agreement with clinicians most clearly for triage recommendation appropriateness and question sufficiency.

For patient interaction dimensions, alignment with non-clinician ratings was limited, with mean absolute Spearman correlations of 0.087 for PLT and 0.120 for rPLT in reference-free mode; for rPLT, alignment increased to 0.176 under reference-based judging. Under reference-free judging, the strongest rPLT correlations were clarity ($\rho = 0.277$), medical jargon ($\rho = -0.128$), and relevance ($\rho = 0.126$), while coherence ($\rho = 0.023$) and fluency ($\rho = 0.045$) showed the weakest alignment. Under reference-based judging, the strongest correlations were relevance ($\rho = 0.249$), fluency ($\rho = 0.206$), and coherence ($\rho = 0.179$), while medical jargon remained weak ($\rho = 0.041$) and clarity ($\rho = -0.204$) remained inconsistent (Figure 4b). Overall, reference-based LLM judging modestly improved alignment with non-clinician ratings, particularly for relevance, fluency, and coherence, but agreement remained weak. In a sensitivity analysis using Claude

Sonnet 4.6 as a judge from a different model family, PLT again received higher scores than rPLT. Mean scores across 12 dimensions in reference-free mode were 4.51 for PLT and 3.71 for rPLT, compared with 4.15 and 3.64 with the GPT-5.2 judge, and PLT was scored higher than rPLT in 38 of 40 cases ($p < 0.001$). The direction of the difference between tools agreed between the two judges in 11 of 12 dimensions; medical jargon was the only dimension scored higher for rPLT by both judges. Alignment of the Claude judge with clinician ratings was similar to that of the GPT-5.2 judge (mean absolute Spearman correlation, 0.193 for PLT and 0.230 for rPLT in reference-free mode and 0.254 for rPLT in reference-based mode, compared with 0.190, 0.186, and 0.223), and alignment with non-clinician ratings remained weak for both judges (Supplementary Information G).

## Simulation evaluation

Simulation evaluation assessed robustness to rare diseases and must-not-miss diagnoses as a descriptive stress test. It included 5 thrombotic thrombocytopenic purpura (TTP) cases, 4 myasthenia gravis (MG) exacerbation cases, and 4 cases each of stroke, pulmonary embolism (PE), and myocardial infarction (MI). Each case was run 20 times per tool. Case-level ESI performance was summarized using the modal prediction across runs, while diagnostic recognition was assessed by the frequency with which the target condition appeared as the primary diagnosis.

For rare diseases, both tools achieved exact modal ESI agreement in 3/5 TTP cases and 4/4 MG exacerbation cases. In the remaining two TTP cases, PLT undertriaged both by one ESI level, whereas rPLT undertriaged one and overtriaged one by one level. Despite similar ESI performance, diagnostic recognition differed. PLT identified MG exacerbation as the primary diagnosis in 45/80 outputs and TTP in 6/100 outputs. rPLT identified neither condition as the primary diagnosis, although it suggested a non-specific neurological or neuromuscular deficit-related diagnosis in 20/80 MG outputs.

For must-not-miss diagnoses, exact modal ESI agreement was observed in 3/4 stroke, 1/4 PE, and 2/4 MI cases for PLT, compared with 4/4, 1/4, and 3/4 cases, respectively, for rPLT. Across these 12 cases, PLT undertriaged two cases by one ESI level and one case by two levels, and overtriaged three cases by one level. rPLT undertriaged one case and overtriaged three cases, all by one level.

PLT more frequently identified the must-not-miss conditions as the primary diagnosis, recognizing stroke in 63/80 outputs, PE in 44/80, and MI in 58/80, compared with 47/80, 21/80, and 40/80 outputs, respectively, for rPLT. Some failures of target diagnosis recognition by rPLT reflected broad diagnostic labels rather than necessarily incorrect diagnoses. For example, in 24/80 PE outputs, rPLT reported "cardiopulmonary disease" without explicitly naming PE. Such labels were insufficiently specific to count as recognition of the target condition.

## DISCUSSION

In this study, we developed SAFE-Triage, a multidimensional framework for evaluating patient-facing emergency triage tools, and applied it to compare PLT and rPLT. The framework identified clinically relevant trade-offs across clinical performance, patient-facing communication, and clinician-facing summary quality that single-metric evaluation would have missed.

The comparison between PLT and rPLT suggests that replacing LLM-based acuity assignment with a statistical model and rule-structured questioning introduced a trade-off between output standardization and clinical performance, rather than uniformly improving triage. PLT achieved higher exact ESI accuracy, lower undertriage, higher acceptable accuracy, and higher clinician ratings across all dimensions. For patient-facing triage tools, false-negative errors are particularly important because undertriage may falsely reassure higher-acuity patients and delay appropriate care. The lower undertriage rate with PLT is therefore especially relevant from a patient safety perspective. In contrast, rPLT produced shorter, more readable outputs and more stable ESI predictions, reflecting deterministic acuity estimation by the statistical model and the more standardized question structure. However, greater stability should not be interpreted as greater clinical correctness. One possible explanation is that rPLT's predefined question pathway and binary representation of symptoms limited the clinical information available for acuity estimation. Because the two tools differed in acuity model, question structure, and underlying LLM (GPT-5.2 for PLT; GPT-4o for most text-generating steps of rPLT), differences between them cannot be attributed to any single component; the findings suggest that a model-based, rule-structured approach may improve conciseness, readability, structure, and consistency without necessarily improving diagnostic reasoning or triage accuracy.

Across both PLT and rPLT, the strongest performance was observed in patient-facing interaction and summary-related dimensions, while the lowest clinician-rated dimensions were question sufficiency, diagnosis accuracy, and triage recommendation appropriateness. This pattern suggests that LLM-based triage approaches can produce understandable patient interactions and usable clinician-facing summaries, but remain less reliable on the clinical tasks most directly tied to patient safety. Simulations of rare diseases and must-not-miss conditions further highlighted these limitations. PLT more frequently identified the target conditions as the primary diagnosis, whereas rPLT achieved exact modal ESI agreement in more must-not-miss cases, illustrating that diagnostic recognition and acuity classification do not necessarily align. TTP recognition remained limited for both tools. This may partly reflect limited elicitation of bleeding-related symptoms, but should be interpreted cautiously because TTP diagnosis usually depends on laboratory evidence, including thrombocytopenia, microangiopathic hemolytic anemia, and schistocytes, and is rarely established with confidence at the initial triage encounter.[33]

Clinician agreement was moderate for exact scores but increased substantially when ratings within one point were considered equivalent, suggesting that most disagreements reflected minor differences in scoring rather than major differences in interpretation. Agreement between clinician and non-clinician ratings was weak for patient interaction dimensions ($\rho = 0.133$ for PLT; $\rho = -0.121$ for rPLT), indicating that clinician assessment may not fully capture how users perceive patient-facing interactions. However, the non-clinician raters were young, healthy graduate students without chronic medical conditions, so future evaluations should include more representative users with varied age, comorbidity, and health literacy profiles.

The cross-module findings show that scalable evaluation methods captured some dimensions better than others. Reference-free LLM judging rated both tools more strictly than clinicians and showed limited agreement with clinician ratings, although agreement improved for rPLT when clinician-generated references were added, particularly for diagnosis accuracy and triage recommendation appropriateness. Automated metrics were most informative for reference-sensitive semantic dimensions such as relevance and completeness, while clarity and fluency showed weak or inconsistent associations with clinician ratings. However, these metrics capture textual properties and are not validated measures of clinical or communication quality. Automated metrics and LLM-as-judge evaluation may therefore support scalable screening, but require validation against human clinical assessment and should not replace clinician review.

The practical value of SAFE-Triage lies in identifying where triage tools may require oversight, pose safety concerns, or introduce inefficiencies across multiple dimensions beyond acuity accuracy alone, with different modules useful at different development stages. Early on, large-scale automated evaluation can identify broad accuracy and undertriage problems across many cases. Closer to deployment, LLM-as-judge evaluation and simulation testing may help identify issues beyond triage accuracy, including question sufficiency, communication quality, robustness to rare disease presentations, and clinician-facing summary quality. Human evaluation remains essential as a clinical oversight layer, supported by the finding that neither automated metrics nor LLM-as-judge evaluation fully agreed with clinician assessment. From a deployment standpoint, the relevant risks for patient-facing triage tools are measurable: unsafe undertriage, unnecessary overtriage, insufficient information gathering, poor communication, user overreliance, and unclear points of clinician oversight. A tool that performs well on retrospective accuracy may still be unsuitable for deployment if it is difficult to integrate into workflow, too burdensome for patients, or not trusted by clinicians.

We also conducted two supplementary analyses. First, free-text chief complaints were mapped to ICPC-2 codes to assess whether standardized complaint representation could improve reproducibility.[34] However, because ICPC-2 was developed primarily for primary care, this coding reduced the granularity of heterogeneous ED symptom presentations and was therefore not used in the primary analysis. Second, downstream clinical outcomes were explored as an alternative reference for validating initial triage decisions. These outcomes were not adopted as primary endpoints because they do not directly measure acuity at presentation: patients may be appropriately classified as high acuity based on potentially life-threatening symptoms yet avoid ICU admission or death following timely assessment and treatment. Hospitalization, ICU admission, and mortality are also influenced by subsequent clinical management, treatment response, comorbidity, and local admission practices. Details are reported in Supplementary Information A.

This study has several limitations. First, the evaluation used retrospective MIMIC-IV-derived vignettes rather than live patient interactions; consequently, we did not assess usability or safety for patients unable to communicate through text, including those with limited literacy, language barriers, or acute or chronic illness impairing communication. Although HPI-derived vignettes approximated information available at first contact, clinical notes are not equivalent to real-time patient self-report and may reflect physician interpretation after initial assessment. In addition, in the simulated interactions, rPLT's follow-up questions were answered only as yes or no, and symptoms that were not documented in the history of present illness were answered as no. Missing information was therefore represented as a negative response; across the repeated runs of the 40 vignettes, 78% of simulated rPLT responses were negative, and the proportion reflecting undocumented rather than explicitly denied symptoms could not be determined. PLT's simulated patient could instead state that information was not known, so the two tools did not receive responses in the same format, which may have contributed to the differences between tools in the 40-vignette evaluation. The 10,030-case acuity analysis did not involve simulated patient interaction, although the symptom indicators used by rPLT in that analysis likewise coded undocumented symptoms as absent. The simulated patient could not provide uncertain or 'not sure' responses to rPLT, limiting generalizability to real patients who may give uncertain, incomplete, or ambiguous answers. The 40 vignettes covered nine common complaint categories and did not include trauma-related presentations or most acute neurological presentations, so findings on interaction and summary quality cannot be generalized to those presentations. In the 10,030-case acuity analysis, which was not restricted by chief complaint, undertriage was higher for acute neurological complaints with both tools (Supplementary Information F). The recorded ESI level is an in-ED assessment made by triage nurses with access to vital signs and direct observation, whereas neither tool received this information. Disagreement with the recorded ESI level therefore does not necessarily indicate tool error, and no expert reference standard based solely on information available before ED arrival was used; the clinician ratings of recommendation appropriateness in the 40 vignettes provide only a limited assessment of the patient-facing recommendations. In addition, all included encounters resulted in hospital admission, and only 43 of the 10,030 encounters had a reference ESI level of 4 or 5. The acuity analysis therefore addresses discrimination among higher-acuity presentations and does not establish whether the tools can identify patients who could safely be directed to primary care or self-care. Second, all cases were derived from MIMIC-IV, limiting generalizability across hospitals, healthcare systems, documentation styles, and patient populations; the dataset also contains relatively few ESI 4 and 5 cases, does not include patients younger than 18 years and does not capture lower-acuity patients who do not present to the ED. Third, automated diagnosis evaluation was not included because MIMIC-IV diagnosis labels may reflect conditions identified later during admission, and ESI was used as a practical but imperfect acuity reference. Although ESI is widely established, it is not globally adopted and other triage scales may address different operational or clinical priorities. Fourth, PLT, its patient simulator, and the LLM-judge were all based on GPT-5.2, and the text-generating steps of rPLT used GPT-4o or GPT-5.2 depending on the experiment; all were from one model family, limiting generalizability, and the difference in model version is a potential confounder of comparisons of text quality between tools. The sensitivity analysis with a non-GPT judge addressed dependence on the judge model only; the evaluated tools and patient simulators were not varied. In addition, rPLT acuity performance in the 10,030-case analysis was estimated on the encounters used to fit the model and is therefore likely optimistic. The PLT patient simulator produced brief and direct responses (median 13 words) and expressed uncertainty in 0.8% of responses across the repeated runs of the 40 vignettes; the findings therefore reflect a cooperative and articulate simulated patient and favorable conditions for information gathering. In addition, because ESI is typically applied by ED nurses, human evaluation by two critical care clinicians may have introduced specialty-related bias. In a supplementary review, two emergency physicians with access to the original scores scored PLT higher than rPLT on all dimensions but assigned lower scores for prudence and user trust, and agreement between them was lower than between the critical care clinicians; this review was not blinded. Human evaluation was limited to 40 cases and two raters in each group, and the non-clinician raters were postgraduate students, who are not representative of the intended patient population. Finally, clinician raters were not blinded to tool identity, which may have introduced expectation or halo effects; the 40-vignette scorecard sample limited power across 13 dimensions and two tools; and the simulation module covered five conditions with four or five cases each and did not include other time-sensitive conditions such as sepsis, ectopic pregnancy, meningitis, or aortic dissection; repeated runs characterize output variability but do not increase the number of independent clinical presentations. Several future directions emerge from this work. A natural next step is to apply SAFE-Triage to larger, external, and clinically diverse datasets, spanning different hospitals and healthcare systems, varied documentation styles and patient populations, lower-acuity presentations, and rare or high-risk diseases that are underrepresented in MIMIC-IV. Moving beyond retrospective vignettes toward prospective evaluation with live or simulated patient interactions will be important for capturing how patients describe symptoms in their own words, including the ways they may omit, misunderstand, or misprioritize information. Future work should also evaluate alternative interaction modes, including voice- or video-based triage interfaces, for patients who cannot reliably use text-based chat. Deeper investigation of the simulation module will also be valuable, particularly to characterize repeated-output stability across triage-related output domains, where stochastic variation may behave differently. Given the limited agreement between scalable evaluation methods and clinician ratings, a further direction is to delineate when automated metrics and LLM-judge evaluation can be relied upon, when clinician-generated reference outputs meaningfully improve alignment, and which dimensions continue to require human review. Translating SAFE-Triage toward deployment will require studies of usability, workflow integration, patient and clinician acceptance, clinician oversight needs, and the broader set of implementation stakeholders involved in emergency department operations, governance, and resource planning. Finally, the framework should evolve alongside emerging AI-enabled triage workflows, particularly as ambient digital scribe systems transition from documentation support toward real-time clinical decision support during or immediately after patient intake.[35]

Overall, SAFE-Triage identified clinically meaningful trade-offs across clinical performance, patient-facing communication, and clinician-facing summary quality. The framework showed that a triage tool can perform well in one domain while remaining weaker in another, and that scalable evaluation methods are useful for breadth and reproducibility but cannot replace human clinical review. In clinical practice, SAFE-Triage may support pre-deployment refinement and ongoing monitoring of patient-facing triage AI tools by helping clinical teams determine where a tool is efficient, where it needs improvement, and where clinician oversight is needed.

# METHODS

SAFE-Triage was adapted from the SCRIBE framework, retaining its four-module structure of human evaluation, auto evaluation, LLM-as-judge evaluation, and simulation testing, and refocusing it on three triage-specific domains: clinical performance, patient interaction quality, and clinician-facing summary quality.[27,28] These domains were selected because patient-facing triage tools must simultaneously reason clinically, communicate clearly with patients, and generate outputs that support downstream clinician review; evaluating any domain in isolation may miss important failure modes. Human review captures clinical nuance, automated and LLM-judge methods support scalable and reproducible assessment, and simulation testing examines tool behavior under challenging inputs. The Methods describe the implementation details for the two triage tools, MIMIC-IV vignette construction, and each evaluation module (Figure 1).

## Triage AI tools

To illustrate the SAFE-Triage framework, we evaluated two approaches to patient-facing triage: PLT, in which an LLM performs the full triage task, and rPLT, in which a statistical acuity model and rule-structured questioning are combined with LLM-generated text. This comparison was designed to assess how the two approaches differed in performance across the SAFE-Triage framework. Both tools were evaluated for the same intended use: helping patients assess symptoms outside the hospital setting and guiding them toward the appropriate level of care.

PLT used ChatGPT 5.2,[32] which was the most recent GPT model available when outputs were generated, and was configured through predefined structured prompts. These prompts instructed the model to perform the triage task and produce the predefined output format without additional rule-based constraints on its clinical reasoning or knowledge use. PLT was run in zero-shot mode with no in-context examples provided. Model temperature and additional parameters used are available at https://github.com/AutoEvaluation/TriageAIEval and https://github.com/AutoEvaluation/TriageAIDev.

rPLT combined a statistical acuity model with LLM-generated patient-facing text. Acuity was estimated in two stages from structured variables derived from MIMIC-IV. In the first stage, a multinomial logistic regression estimated a baseline expected ESI level from age, sex, history of hypertension and diabetes, and nine chief complaint indicators. In the second stage, a regression tree fitted to the residuals of the first stage adjusted this estimate using binary symptom indicators derived from the chief complaint and the history of present illness through rule-based methods, natural language processing (NLP), and LLM-assisted extraction (Supplementary Information B and C); the adjusted estimate was rounded to the nearest ESI level. Pain score and race were not used as predictors, and recorded ESI version 4 levels were used only as reference labels for model development and evaluation. The splitting variables of the regression tree defined the sequence of follow-up questions. In the patient-facing interface, rPLT accepted a free-text chief complaint and then asked yes/no follow-up questions corresponding to these variables, supplemented by a small number of LLM-generated complaint-specific yes/no questions. An LLM was used to rephrase questions in plain language, to select the patient-facing care recommendation corresponding to the predicted ESI level, and to generate the diagnosis, differential diagnosis, and SBAR summary. GPT-4o was used for LLM-assisted feature extraction and for these steps in the repeated-output stability runs and the rare-disease simulations. In the outputs rated in the human evaluation, the SBAR summaries and remaining steps were generated with GPT-5.2. In the must-not-miss simulations, GPT-5.2 was used for all rPLT LLM steps because GPT-4o was no longer available. The predicted ESI level in rPLT was therefore determined by the statistical model and not by the LLM.

Both PLT and rPLT received the patient's chief complaint and could ask targeted follow-up questions. In the vignette tests, the history of present illness provided to both tools and their patient simulators was truncated at the first documentation of ED assessment or management, limiting inputs to information patients could report before arrival, such as symptoms and self-measured fever, and excluding clinician-measured vital signs, physical examination findings, and test results. For each structured case, an LLM patient simulator (GPT-5.2 for PLT; GPT-4o for rPLT, or GPT-5.2 in the must-not-miss simulations) responded using only information available in the corresponding case record. The PLT simulator was instructed to answer in the first person in lay wording, to disclose information progressively without volunteering details that were not asked for, and to state that information was not known when it was not supported by the record. For rPLT, responses were restricted to yes or no, and symptoms not documented in the history of present illness were answered as no; for PLT, the simulated patient could state that information was not known. In rPLT, the predicted ESI level was computed from the structured case variables and did not depend on the simulated responses. After the interaction, both tools generated a diagnosis and differential diagnosis, a patient-facing care recommendation, and a clinician-facing SBAR summary with a predicted ESI level.

The patient-facing recommendations generated by both tools were mapped to five predefined care categories informed by prior patient-facing symptom checker studies.[36,37,38] The categories were ordered by urgency and included call an ambulance, seek emergency care, seek urgent medical care, seek primary care non-urgent, and self-care (Table 2).

## Clinical vignette development

Clinical cases were derived from MIMIC-IV, a deidentified electronic health record database containing ED encounters with structured triage information, diagnoses, chief complaints, and clinical notes. This made MIMIC-IV suitable for constructing ED presentations and comparing tool outputs with recorded ESI acuity levels. All included encounters were ED visits that resulted in hospital admission. To avoid data leakage between rPLT development and framework evaluation, all encounters used for the final evaluation vignettes were excluded from the data used to derive and validate the rule-augmented triage workflow.

The 40 clinical vignettes were selected across predefined, commonly encountered chief complaint categories and ESI levels to demonstrate the SAFE-Triage evaluation framework. Target numbers within each complaint category were informed by their relative frequency in ED presentations.[4,39,40]

NLP-based methods were used to identify eligible MIMIC-IV encounters, followed by random sampling within each category. Trauma presentations were outside the selected case spectrum because they may require different assessment pathways or trauma-specific questioning. Continuous vignette characteristics were summarized as medians and interquartile ranges, because graphical examination showed that age was approximately normally distributed (Supplementary Information E). Shapiro–Wilk tests were consistent with this assessment (age, $P = 0.021$). Details can be found in Supplementary Information C.

Cases were included only when complete data were available, including ESI level, chief complaint, diagnosis, age, and sex. Patients also had to be formally triaged upon arrival. In addition, only cases with a history of present illness section longer than 80 words were included to ensure sufficient symptom detail for vignette construction. To ensure consistent documentation, only encounters between 2016 and 2019 were included, corresponding to the period after implementation of a new electronic health record system and before workflow disruptions associated with the COVID-19 pandemic. Cases were excluded if they appeared to bypass standard triage because of immediate emergency intervention or insufficient triage information at arrival, such as CPR, severe active bleeding, or emergency airway management.

To improve representation across the full triage spectrum, additional lower-acuity cases were manually included, particularly ESI 4 and ESI 5 presentations, which were relatively infrequent in MIMIC-IV. The resulting case set was then transformed into structured clinical vignettes describing standardized patient presentations, including demographics, chief complaint, symptom description, and relevant clinical context. All vignettes were subsequently reviewed by an emergency medicine clinician (N.P.) and two triage nurses (D.P. and K.N.) to ensure clinical plausibility, diversity in presentation and acuity, and consistency with realistic ED scenarios.

## Human evaluation

Human evaluation was conducted using the annotated scorecard provided in Supplementary Information D, including domain definitions, rating anchors, and scoring instructions. The clinical performance domain included diagnosis correctness, triage recommendation appropriateness, and whether the tool asked sufficient relevant questions to support clinical safety. The patient interaction and dialogue quality domain included clarity, fluency, coherence, relevance, appropriateness of medical terminology, and user trust. The clinician-facing summary quality domain included completeness, SBAR adherence, clinical prudence, and brevity.

Two critical care clinicians (H.F. and M.Al.) independently rated all triage tool outputs. Before formal evaluation, both clinicians completed a five-case calibration exercise followed by joint review to align interpretation of the scorecard domains and rating anchors. To assess the influence of rater specialty, two emergency physicians (M.B. and M.Ar.) subsequently reviewed all 40 cases for both tools. They had access to the critical care clinicians' scores and assigned their own scores on the same dimensions. Their scores were averaged without adjudication and were used only for this supplementary comparison (Supplementary Information H).

In parallel, two postgraduate students without extended medical knowledge (E.B. and J.C.K.L.) independently rated the patient interaction and dialogue quality domain. Student raters completed the same five-case calibration procedure and were analyzed separately from clinician raters.

Inter-rater agreement was assessed before aggregation. Ratings meeting the pre-specified substantial-disagreement threshold underwent adjudication, after which final human scores were calculated using the adjudicated score for those items and the mean of the two raters otherwise. Scorecard ratings are ordinal and are reported as means across cases as descriptive summaries; no parametric test was applied to them, and their associations were assessed using Spearman correlation.

In addition to assigning scorecard ratings, the two clinicians independently revised rPLT outputs to create clinician-generated reference outputs. These revisions reflected each clinician's preferred version of the triage output based on clinical judgment and included adding missing questions, removing unnecessary content, and rephrasing text when needed. These references were created only for rPLT and were used for reference-based automated metrics and reference-based LLM-judge evaluation.

Human-derived materials were used in two distinct ways. Final clinician scores served as the primary human reference standard for comparing PLT and rPLT across individual scorecard dimensions and grouped domains. Final non-clinician scores were used as a secondary human assessment of the patient interaction and dialogue quality domain. Clinician-generated reference outputs were used separately as text-based reference standards for reference-based auto evaluation and reference-based LLM-judge evaluation. Thus, human input contributed both numeric ratings and edited reference outputs, with the former supporting scorecard-based comparison and the latter supporting reference-based assessment of output similarity and rubric-guided LLM evaluation.

## Auto evaluation

For the triage classification analysis, we randomly selected 10,030 MIMIC-IV ED encounters using the same inclusion and exclusion criteria as the clinical vignettes. This analysis did not involve simulated patient interaction; each tool generated a predicted ESI level directly from the case information. For each encounter, the predicted ESI level was compared with the reference ESI level. The recorded ESI level was assigned by ED triage nurses with access to vital signs and direct observation, which were not available to either tool, and was used as a retrospective benchmark. Predictions less urgent or more urgent than the recorded level are termed undertriage and overtriage, respectively, without implying adjudicated clinical error. Paired differences between PLT and rPLT were assessed using two-sided McNemar tests with 95% confidence intervals, because both tools were applied to the same encounters and the outcomes were binary. Because encounters were not restricted by chief complaint, exploratory post hoc subgroup analyses were performed for trauma-related complaints, acute neurological complaints, and complaints within and outside the nine vignette categories, identified from the free-text chief complaint using keyword rules (Supplementary Information F). Two versions of the rPLT acuity model were used. For the 40 vignettes, the model combined ESI levels 4 and 5 and was developed on encounters that excluded the vignettes. For the 10,030-case analysis, a five-level version was fitted to the same 10,030 encounters; in an internal 20% hold-out, its exact accuracy was 39.6%. Automated diagnosis analysis was not performed because MIMIC-IV diagnosis labels may include conditions identified during admission rather than at the initial ED encounter.

To assess repeated-output stability, each of the 40 clinical vignettes was run 20 times for both PLT and rPLT using identical parameters. Stability was assessed for ESI prediction, patient-facing triage recommendation, primary diagnosis, and combined primary and differential diagnosis outputs. For ESI prediction and patient-facing triage recommendation, we calculated modal-category consistency, defined as the proportion of repeated runs that produced the most frequently observed category for the same case. This approach adapts majority-agreement methods used to quantify agreement across repeated labels.[41,42] Diagnostic stability was assessed using cosine similarity for the primary diagnosis and for the combined primary and differential diagnosis outputs. Differences in categorical stability were compared using two-sided exact paired sign tests, while diagnostic cosine similarity metrics were compared using two-sided paired Wilcoxon signed-rank tests. These nonparametric paired tests were used because both stability measures are bounded and no distributional form was assumed for their paired differences.

Automated metrics were applied to scorecard dimensions for which quantitative measures could be defined. These metrics measure textual properties and were used as proxies for the corresponding dimensions, not as validated measures of clinical or communication quality. Reference-free metrics were calculated for both PLT and rPLT using only the original tool outputs and without clinician-generated gold standard reference, including perplexity for fluency, Flesch Reading Ease for clarity, Medical Term Recall for medical terminology, and compression ratio for brevity.[43–46] Reference-based evaluation was performed only for rPLT and used clinician-generated gold standards, created by having both clinicians independently revise rPLT outputs to match their clinical judgment by adding missing questions, removing unnecessary content, and rephrasing text where needed. Because the references were derived from rPLT outputs, similarity to them partly reflects the extent of clinician editing and not agreement with an independent standard. Metrics were calculated against each clinician-generated gold standard reference and averaged as the final auto evaluation score; these included ROUGE for completeness, BERTScore for relevance, and BLEU/METEOR for similarity to reference texts, while dimensions without a robust automated metric, such as coherence, were evaluated through the human and LLM-judge modules (Table 2).[47–50] Spearman correlations were then calculated between scorecard-aligned automated metrics and the corresponding clinician ratings.

## LLM-as-judge evaluation

To complement human and auto evaluation, we incorporated an LLM-judge using GPT-5.2 to score triage AI outputs according to the same rubric as the human evaluation scorecard. The LLM-judge was designed to mimic structured human assessment by assigning scores on the same 1–6 scale for each predefined evaluation dimension, thereby enabling scalable and rubric-aligned assessment across cases.

The LLM-judge was prompted with the complete evaluation score rubric used by the clinicians and instructed to score each case strictly according to the rubric definitions and anchors. A fixed prompt template was used across all cases, with role-specific framing by domain: clinician perspective for clinical quality and clinician-facing summary domains, and patient perspective for patient interaction and dialogue quality domain (Supplementary Information B).

Two LLM judging modes were evaluated. Reference-free judging used the same source case materials, triage conversation, and original tool output as the human raters, and scored both PLT and rPLT using the same evaluation rubric. Reference-based judging additionally included the clinician-generated rPLT reference output as a scoring reference and was therefore performed only for rPLT.

To evaluate alignment with human assessment, Spearman correlation coefficients were calculated between LLM-judged scores and clinician ratings for each evaluation dimension.

To reduce stochastic variation and assess scoring stability, the LLM-judge was run 7 times per case and evaluation dimension under identical prompting conditions, and the median score across runs was retained as the final LLM-judge score. To assess scoring stability, the standard deviation across runs was computed for each case–dimension combination, and dimensions with a standard deviation greater than 0.5 were flagged as unstable and considered to require prompt refinement. Scoring stability scores are provided in Supplementary Information B. To assess dependence on the judge model, the LLM-judge evaluation was repeated with Claude Sonnet 4.6 (Anthropic; model identifier claude-sonnet-4-6) using the same rubric, with seven runs per case and judging mode and the median across runs retained.[51] The maximum output length was set to 4,096 tokens, and the sampling temperature was left at the service default. In reference-based mode, the Claude judge received both clinician-generated reference outputs together, whereas GPT-5.2 scores were obtained against each reference separately and averaged. User trust was not scored in this analysis, which therefore covered 12 dimensions. Scores from the two judges were compared descriptively, and the difference between tools under the Claude judge was assessed using a two-sided exact paired sign test on case-level mean scores (Supplementary Information G).

## Simulation evaluation

Simulation evaluation was used as a descriptive stress test to examine robustness in rare disease and must-not-miss presentations. Cases were randomly selected from MIMIC-IV using the same inclusion and exclusion criteria as the clinical vignettes. Rare disease testing included 5 TTP cases and 4 MG exacerbation cases. These conditions can present with non-specific symptoms but require timely recognition and appropriate triage to prevent adverse outcomes.[52–55] Their selection was informed by a clinical expert panel, whose participating physicians prioritized conditions likely to be encountered in emergency practice from a list of rare and uncommon diseases. Must-not-miss testing assessed performance in additional time-sensitive conditions and included 12 cases: 4 acute ischemic stroke cases, 4 acute myocardial infarction cases, and 4 pulmonary embolism cases.

Each case was run 20 times for both PLT and rPLT using identical parameters. ESI accuracy was assessed by comparing the modal prediction across runs with the recorded ESI level, with discrepancies described as undertriage or overtriage. Diagnostic recognition was assessed by the frequency with which the target condition appeared as the primary diagnosis. Findings were summarized descriptively.

## Ethics approval

This study used deidentified MIMIC-IV data obtained through PhysioNet under the applicable Data Use Agreement. The collection of patient information and creation of the MIMIC-IV research resource were reviewed and approved by the Institutional Review Board of Beth Israel Deaconess Medical Center (IRB protocol no. 2001P001699), which granted a waiver of informed consent and approved the data-sharing initiative. The present study involved only secondary analysis of deidentified, credentialed-access data, and no identifiable patient information was accessed. All analyses were conducted in accordance with the Declaration of Helsinki, the MIMIC-IV Data Use Agreement, and applicable ethical guidelines.

## DATA AVAILABILITY

The datasets analyzed during the current study are not publicly available due to privacy restrictions and the terms of the PhysioNet Credentialed Health Data Use Agreement.[31] MIMIC-IV is available through PhysioNet to credentialed users who meet the required access criteria. Human evaluation scorecard, rubric, and summary statistics are available in the supplementary materials.

## CODE AVAILABILITY

Code used for case extraction, NLP processing, PLT and rPLT implementation, and evaluation analyses is available at https://github.com/AutoEvaluation/TriageAIEval and https://github.com/AutoEvaluation/TriageAIDev.

## ACKNOWLEDGEMENTS

N.E.-Z. was supported by a grant from The Duke Endowment for her efforts. We thank Christopher J. Lindsell for his valuable comments and suggestions on this work.

# CONFLICT OF INTEREST

A.I.W. holds equity and management positions in Ataia Medical, AI Wong Consulting, LLC, and Synappsed, LLC, and serves as a consultant for Merck Sharp & Dohme, LLC. The other authors do not have a competing interest.

# AUTHOR CONTRIBUTIONS

Conceptualization: R.H., C.H.; Methodology: R.H., Z.Q., J.W., E.B., Y.Y., H.W., R.Y., N.L., C.H.; Vignette creation: R.H., Z.Q., J.W., N.R.P., D.P., K.N.; Human evaluation raters: H.F., M.Al., E.B., J.C.K.L.; Emergency physician review: M.B., M.Ar.; Data collection and processing: R.H., Z.Q., J.W., E.B.; Writing: R.H., H.F., M.Al., Z.Q., J.W., E.B., J.C.K.L., H.W., K.Y., N.E.-Z., J.R.V., A.I.W., A.C., C.H.; Guarantor: C.H.; All authors read and approved the final manuscript.

## TABLES AND FIGURES

Figure 1. Overall study workflow. Three case sets were derived from MIMIC-IV emergency department records: 40 clinical vignettes, 21 simulation cases (rare disease and must-not-miss conditions), and 10,030 encounters. For the vignettes and simulation cases, each tool conducted a simulated triage interaction, PLT with an LLM patient simulator that returned free-text responses and rPLT with yes/no responses derived from the case record. For the 10,030 encounters, both tools predicted the ESI level directly from case information, without simulated interaction. In PLT, the LLM performed the full triage task; in rPLT, a statistical acuity model assigned the acuity level and an LLM generated the text outputs. Tool outputs were evaluated through four modules (human evaluation, LLM-as-judge, automated evaluation, and simulation testing) across three domains (clinical performance, patient interaction quality, and clinician-facing summary quality). Colors indicate the case set used in each evaluation component. ESI, Emergency Severity Index; LLM, large language model; PLT, prompted LLM triage; rPLT, rule-augmented prompted LLM triage; SBAR, Situation, Background, Assessment, Recommendation.

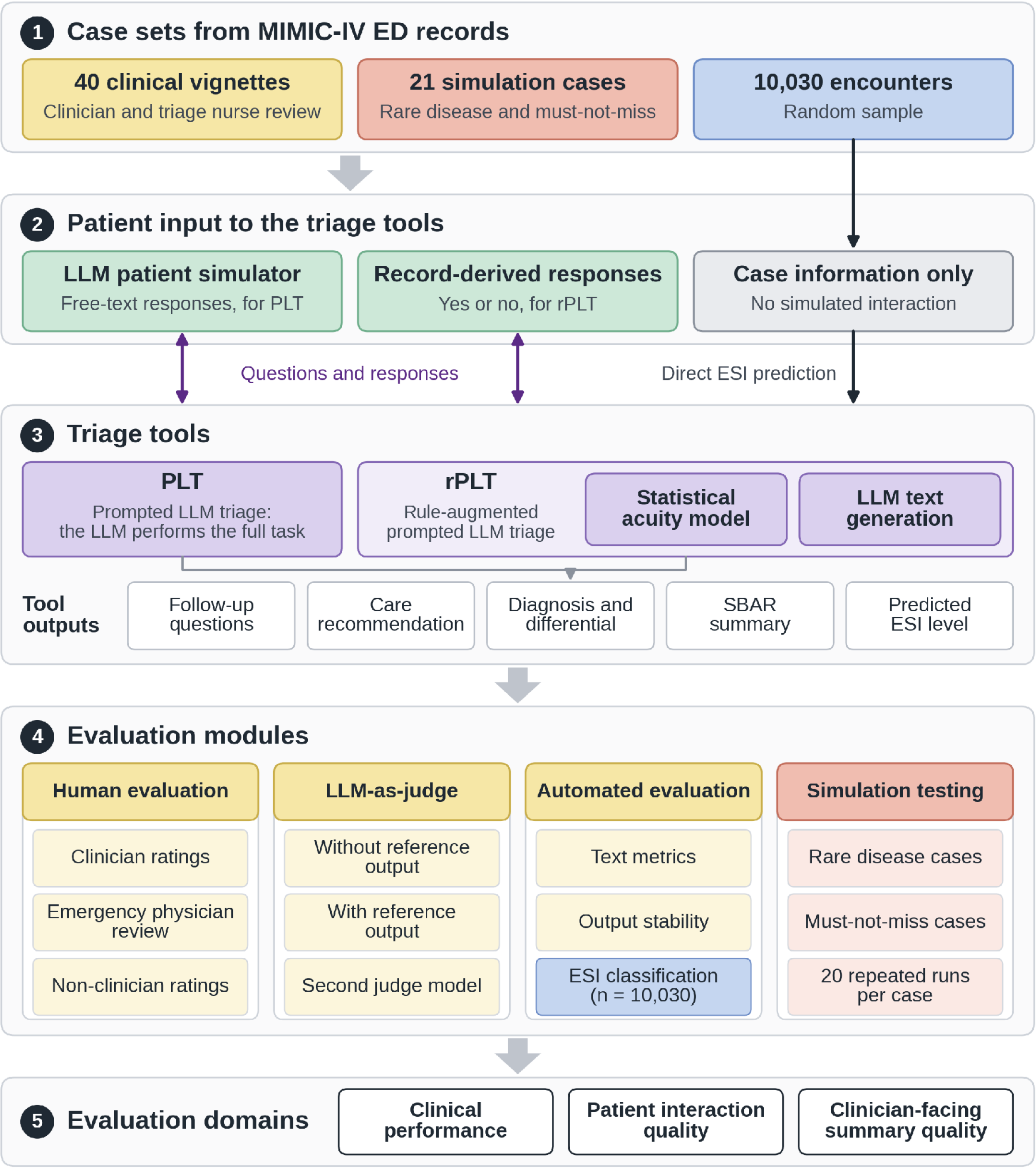
1 Case sets from MIMIC-IV ED records
40 clinical vignettes
Clinician and triage nurse review
21 simulation cases
Rare disease and must-not-miss
10,030 encounters
Random sample
2 Patient input to the triage tools
LLM patient simulator
Free-text responses, for PLT
Record-derived responses
Yes or no, for rPLT
Case information only
No simulated interaction
Questions and responses
Direct ESI prediction
3 Triage tools
PLT
Prompted LLM triage:
the LLM performs the full task
rPLT
Rule-augmented
prompted LLM triage
Statistical acuity model
LLM text generation
Tool outputs
Follow-up questions
Care recommendation
Diagnosis and differential
SBAR summary
Predicted ESI level
4 Evaluation modules
Human evaluation
Clinician ratings
Emergency physician review
Non-clinician ratings
LLM-as-judge
Without reference output
With reference output
Second judge model
Automated evaluation
Text metrics
Output stability
ESI classification (n = 10,030)
Simulation testing
Rare disease cases
Must-not-miss cases
20 repeated runs per case
5 Evaluation domains
Clinical performance
Patient interaction quality
Clinician-facing summary quality

Figure 2. Multi-rater comparison of PLT and rPLT across SAFE-Triage scorecard dimensions. Radar plots show mean scores on the 6-point Likert scale for prompted LLM triage (PLT) and rule-augmented prompted LLM triage (rPLT). Panels show ratings separately by tool and evaluation domain: **(a)** rPLT clinical performance and clinician-facing summary dimensions, rated by clinicians, the reference-free LLM-judge, and the reference-based LLM-judge anchored to clinician-generated rPLT reference outputs; **(b)** rPLT patient interaction and dialogue quality dimensions, rated by clinicians, non-clinician raters, the reference-free LLM-judge, and the reference-based LLM-judge; **(c)** PLT clinical performance and clinician-facing summary dimensions, rated by clinicians and the reference-free LLM-judge; and **(d)** PLT patient interaction and dialogue quality dimensions, rated by clinicians, non-clinician raters, and the reference-free LLM-judge. Reference-based LLM-judge evaluation was performed only for rPLT because clinician-generated reference outputs were created only for rPLT.

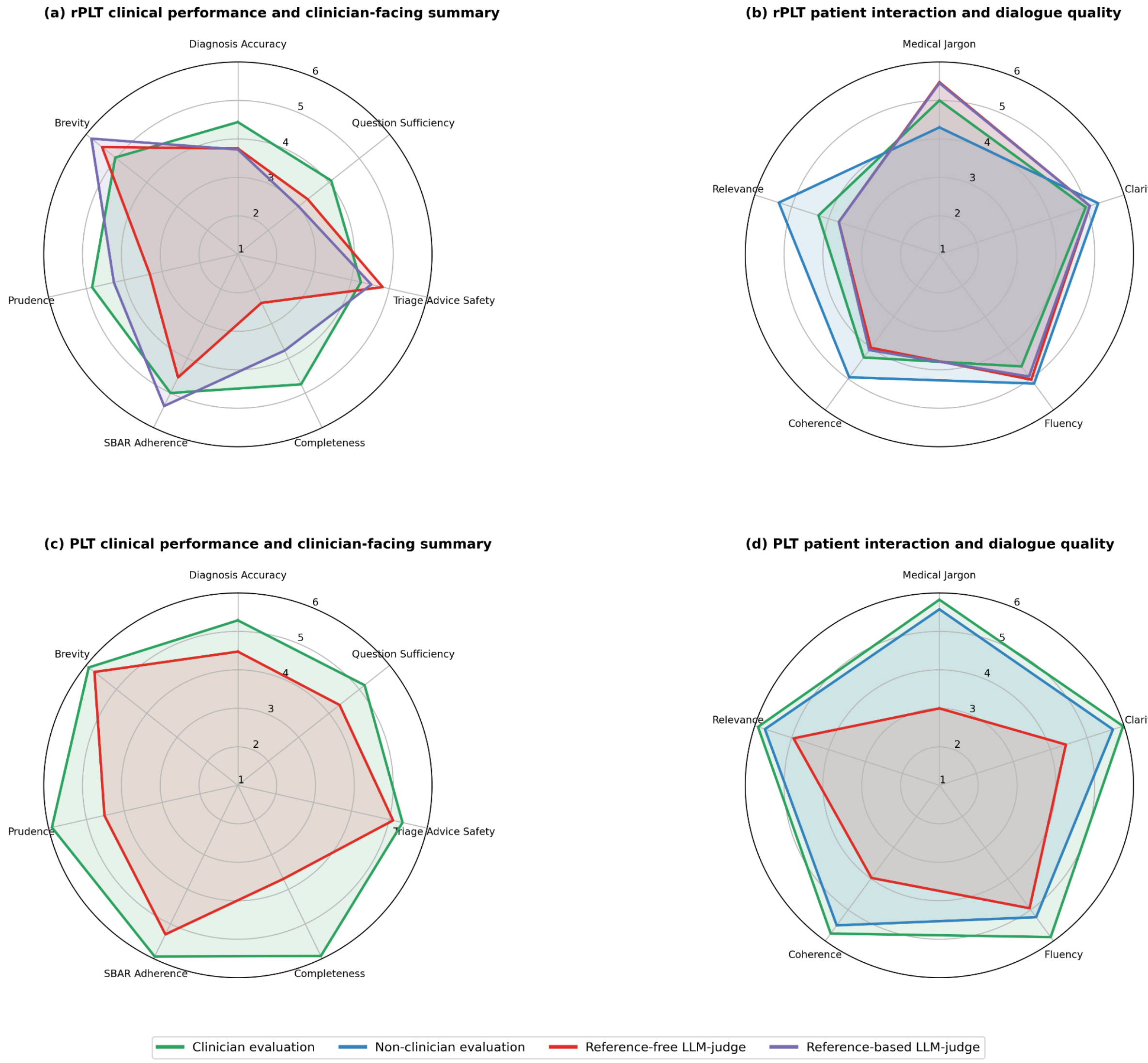

Figure 3. Triage classification performance of PLT and rPLT in the 10,030-case MIMIC-IV acuity analysis. (a) Confusion matrices showing predicted versus reference MIMIC-IV ESI levels for prompted LLM triage (PLT) and rule-augmented prompted LLM triage (rPLT). Darker shading indicates higher case counts, and outlined cells indicate acceptable agreement, defined as exact ESI agreement or prediction of one level more conservative than the reference label. (b) Stacked bar plot showing the proportion of exact, undertriaged, and overtriaged predictions for each tool, stratified by error severity.

(a)

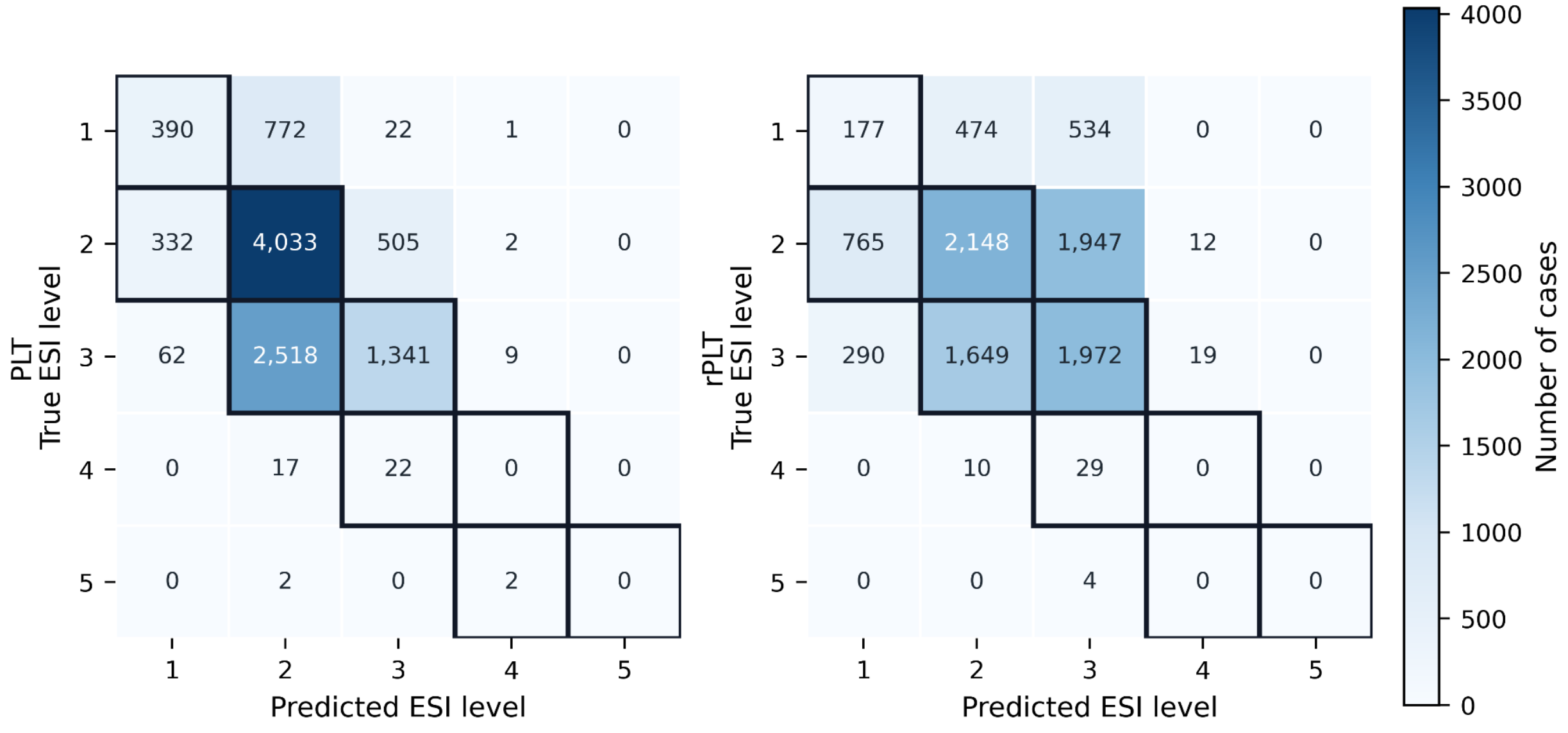

(b)

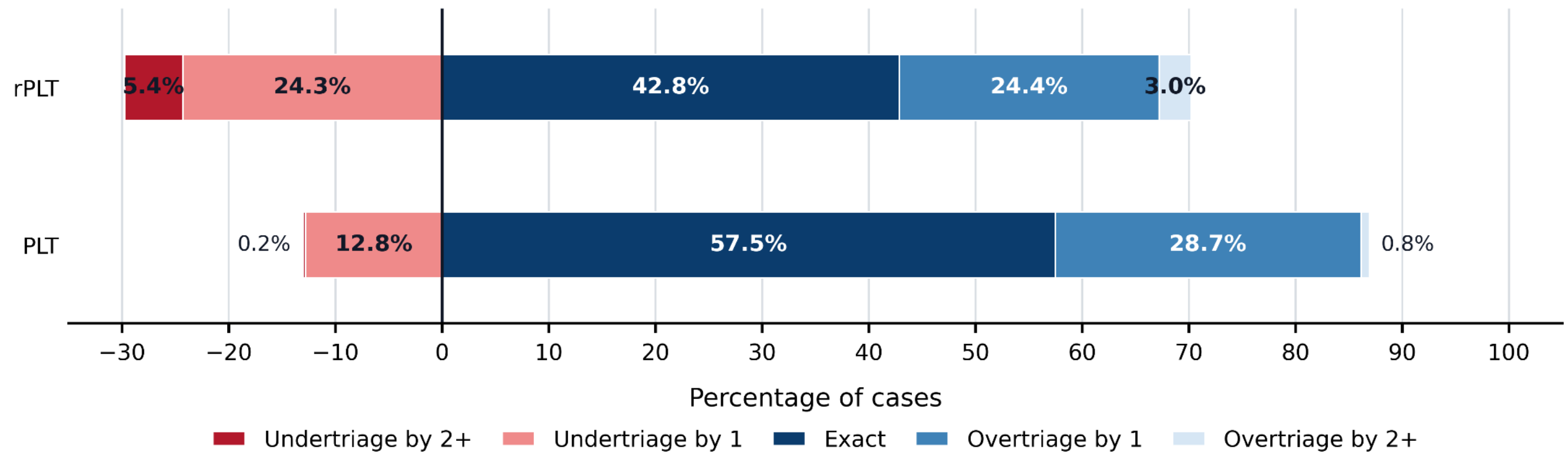

Figure 4. Agreement between evaluation modules and human ratings, with (a) showing clinician ratings and (b) non-clinician ratings, across scorecard dimensions. Heatmap values represent absolute Spearman correlations between human scores and reference-free LLM judging (GPT-5.2), reference-based LLM judging (GPT-5.2) where available, and scorecard-aligned automated metrics for PLT and rPLT.

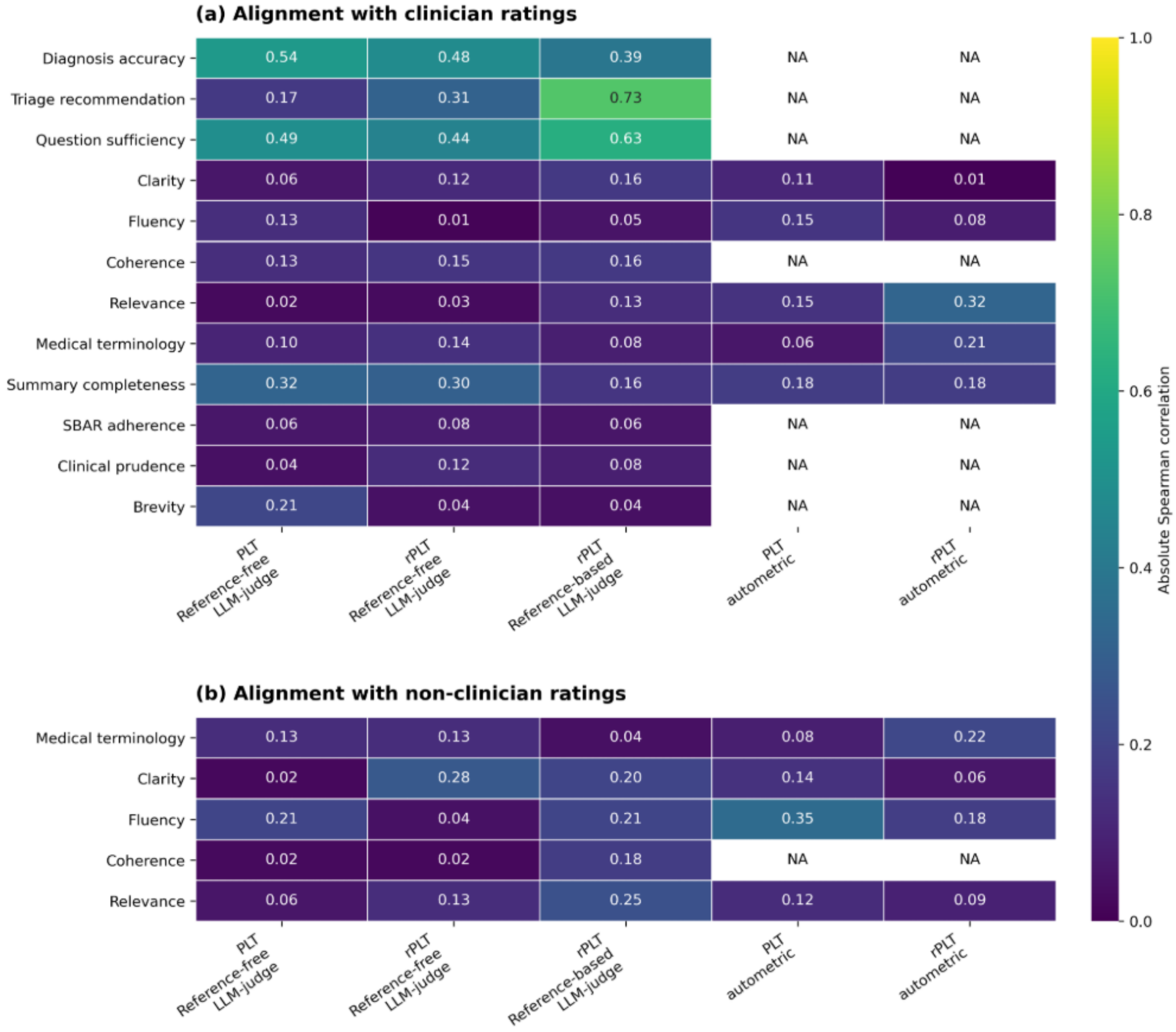

Table 1. Evaluation modules and categories. Reference-based evaluation was not performed for Prompted LLM triage (PLT), since no clinician-generated gold standard was gathered.

| **Triage tool section** | **Evaluation dimension** | **Clinician evaluation** | **Non-clinician evaluation** | **LLM evaluation (clinician)** | **LLM evaluation (patient)** | **Auto evaluation (reference-free)** | **Auto evaluation (reference-based)** | **Simulation evaluation** |
|---|---|---|---|---|---|---|---|---|
| Triage accuracy and output stability | ESI prediction accuracy | | | | | Exact ESI accuracy; Acceptable accuracy; under-, overtriage | | |
| | ESI prediction and triage recommendation stability | | | | | Modal-category consistency | | |
| | Diagnosis and differential stability | | | | | Cosine similarity | | |
| Clinical quality | Triage recommendation appropriateness | X | | X | | | | |
| | Diagnosis accuracy | X | | X | | | | |
| | Question sufficiency | X | | X | | | | |
| | Rare disease and Must-not-miss diagnosis recognition | | | | | | | X |
| Patient Interaction | Medical jargon usage | X | X | | X | | Medical Term Recall | |
| | Clarity | X | X | | X | | Flesch Reading Ease | |
| | Fluency | X | X | | X | | Perplexity | |
| | Coherence | X | X | | X | | | |
| | Relevance | X | X | | X | BERTScore | | |
| | User trust | X | X | | | | | |
| Clinician Summary | Completeness | X | | X | | ROUGE & BLEU/METEOR | | |
| | SBAR Adherence | X | | X | | | | |
| | Prudence | X | | X | | | | |
| | Brevity | X | | X | | | Compression ratio | |

Table 2. Patient-facing triage recommendation categories. Categories are ordered from highest to lowest urgency.

| **Severity rank** | **Patient-facing triage advice** |
|---|---|
| 1 | Call an ambulance |
| 2 | Seek emergency care |
| 3 | Seek urgent medical care |
| 4 | Seek primary care, non-urgent |
| 5 | Self-care |

Table 3. Baseline characteristics of 40 clinical vignettes. Cases selected from MIMIC-IV with HPI ≥ 80 words and full data availability. NRS, Numeric Rating Scale; ESI, Emergency Severity Index.

| Characteristic | N (%), Median (IQR), or Mean ± SD |
|---|---|
| Total | 40 (100%) |
| **Sex** | |
| Female | 20 (50.0%) |
| Male | 20 (50.0%) |
| **Age (years)** | |
| Median (IQR) | 55.5 (33.5–68.0) |
| Range | 20 – 85 |
| 20 – 34 | 12 (30.0%) |
| 35 – 49 | 4 (10.0%) |
| 50 – 64 | 11 (27.5%) |
| 65 – 79 | 9 (22.5%) |
| ≥ 80 | 4 (10.0%) |
| **Chief Complaint** | |
| Abdominal pain | 8 (20.0%) |
| Chest pain | 7 (17.5%) |
| Dyspnea | 6 (15.0%) |
| Fever | 5 (12.5%) |
| Headache | 4 (10.0%) |
| Back pain | 3 (7.5%) |
| Altered mental status | 3 (7.5%) |
| Dizziness / lightheadedness | 2 (5.0%) |
| Vomiting / nausea | 2 (5.0%) |
| **ESI Triage Level** | |
| ESI 1 (Immediate) | 5 (12.5%) |
| ESI 2 (Emergent) | 12 (30.0%) |
| ESI 3 (Urgent) | 15 (37.5%) |
| ESI 4 (Less urgent) | 6 (15.0%) |
| ESI 5 (Non-urgent) | 2 (5.0%) |
| **HPI Word Count (Complaint Summary)** | |
| Median (IQR) | 332 (244.5–371.3) |
| Range | 108 – 585 |

Table 4. Mean scorecard-based ratings after adjudication and LLM-judge evaluation (GPT-5.2). Mean scores are shown for clinician and non-clinician ratings, as well as reference-free and reference-based LLM-judge evaluations. All dimensions were rated on a 1 to 6 Likert scale, with higher scores indicating better performance. Reference-based LLM-judge scores were not available for PLT because no clinician-generated reference output was available for this tool.

| **Evaluation dimension** | **Clinician rating** | | **Non-clinician rating** | | **Reference-free LLM-judge** | | **Reference-based LLM-judge** | |
|---|---|---|---|---|---|---|---|---|
| | **PLT** | **rPLT** | **PLT** | **rPLT** | **PLT** | **rPLT** | **PLT** | **rPLT** |
| Diagnosis accuracy | 5.29 | 4.44 | - | - | 4.47 | 3.75 | - | 3.73 |
| Question sufficiency | 5.18 | 4.08 | - | - | 4.35 | 3.30 | - | 3.00 |
| Triage recommendation appropriateness | 5.35 | 4.25 | - | - | 5.10 | 4.83 | - | 4.53 |
| Medical jargon | 5.83 | 5.00 | 5.58 | 4.30 | 3.00 | 5.47 | - | 5.45 |
| Clarity | 5.98 | 4.96 | 5.70 | 5.30 | 4.42 | 5.08 | - | 5.08 |
| Fluency | 5.88 | 4.60 | 5.24 | 5.15 | 4.95 | 5.03 | - | 4.92 |
| Coherence | 5.76 | 4.31 | 5.50 | 4.95 | 3.98 | 4.00 | - | 4.08 |
| Relevance | 5.91 | 4.28 | 5.72 | 5.35 | 4.95 | 3.73 | - | 3.73 |
| User trust | 5.94 | 4.61 | 5.24 | 3.67 | - | - | - | - |
| Completeness | 5.93 | 4.75 | - | - | 3.70 | 2.40 | - | 3.77 |
| SBAR adherence | 5.94 | 5.00 | - | - | 5.30 | 4.55 | - | 5.38 |
| Prudence | 5.91 | 4.85 | - | - | 4.53 | 3.33 | - | 4.28 |
| Brevity | 5.91 | 5.04 | - | - | 5.72 | 5.47 | - | 5.83 |

**Supplementary Materials of**

# "Beyond Acuity Accuracy: A Multidimensional Framework for Evaluating Patient-Facing Emergency Triage AI Tools"

## Supplementary Information A: Additional results

This Supplementary Information provides results for the 10,030-case MIMIC-IV acuity analysis and detailed pre-adjudication inter-rater agreement results, including exact agreement, within-one-point agreement, and substantial disagreement patterns.

### A.1: Baseline characteristics of 10k cases for the triage classification analysis. Cases selected from MIMIC-IV with the same in- and exclusion criteria as the 40 clinical vignettes.

| Baseline characteristic | Value |
|---|---|
| Total cases | 10,030 |
| Mean age | 63.5 years |
| Median age | 65 years |
| Female | 5,079 (50.6%) |
| Male | 4,951 (49.4%) |
| ESI 1 | 1,185 (11.8%) |
| ESI 2 | 4,872 (48.6%) |
| ESI 3 | 3,930 (39.2%) |
| ESI 4 | 39 (0.4%) |
| ESI 5 | 4 (0.04%) |

**A.2.1: Clinician inter-rater agreement before adjudication for PLT per evaluation dimension.**

Exact and within-one-point agreement between the two clinician raters before adjudication, shown by scorecard dimension and overall.

| Clinician PLT Exact Inter-Rater Agreement (1-6 Scale) | |
|---|---|
| **Category** | **Agreement (%)** |
| Diagnosis Accuracy | 17.5% |
| Question Sufficiency for Clinical Safety | 40.0% |
| Triage Advice / Safety Accuracy | 17.5% |
| Medical Jargon | 75.0% |
| Clarity | 95.0% |
| Fluency | 75.0% |
| Coherence | 52.5% |
| Relevance | 82.5% |
| User Trust | 87.5 |
| Completeness | 85.0 |
| SBAR Adherence | 87.5 |
| Prudence | 85.0% |
| Brevity | 82.5% |
| **Overall** | **67.9%** |

| Clinician PLT Within-one-point Inter-Rater Agreement (±1, 1-6 Scale) | |
|---|---|
| **Category** | **Agreement (%)** |
| Diagnosis Accuracy | 95.0% |
| Question Sufficiency for Clinical Safety | 95.0% |
| Triage Advice / Safety Accuracy | 92.5% |
| Medical Jargon | 100.0% |
| Clarity | 100.0% |
| Fluency | 100.0% |
| Coherence | 100.0% |
| Relevance | 100.0% |
| User Trust | 100.0% |
| Completeness | 100.0% |
| SBAR Adherence | 100.0% |
| Prudence | 97.5% |
| Brevity | 100.0% |
| **Overall** | **98.5%** |

**A.2.2: Clinician inter-rater agreement before adjudication for rPLT per evaluation dimension.**

Exact and within-one-point agreement between the two clinician raters before adjudication, shown by scorecard dimension and overall.

| Clinician rPLT Exact Inter-Rater Agreement (1-6 Scale) | |
|---|---|
| **Category** | **Agreement (%)** |
| Diagnosis Accuracy | 25.0% |
| Question Sufficiency for Clinical Safety | 40.0% |
| Triage Advice / Safety Accuracy | 37.5% |
| Medical Jargon | 90.0% |
| Clarity | 75.0% |
| Fluency | 40.0% |
| Coherence | 30.0% |
| Relevance | 32.5% |
| User Trust | 47.5% |
| Completeness | 60.0% |
| SBAR Adherence | 55.0% |
| Prudence | 40.0% |
| Brevity | 60.0% |
| **Overall** | **48.7%** |

| Clinician rPLT Within-one-pointInter-Rater Agreement (±1, 1-6 Scale) | |
|---|---|
| **Category** | **Agreement (%)** |
| Diagnosis Accuracy | 77.5% |
| Question Sufficiency for Clinical Safety | 90.0% |
| Triage Advice / Safety Accuracy | 87.5% |
| Medical Jargon | 100.0% |
| Clarity | 97.5% |
| Fluency | 95.0% |
| Coherence | 87.5% |
| Relevance | 92.5% |
| User Trust | 80.0% |
| Completeness | 80.0% |
| SBAR Adherence | 95.0% |
| Prudence | 75.0% |
| Brevity | 97.5% |
| **Overall** | **88.8%** |

**A.3.1: Non-clinician inter-rater agreement before adjudication for PLT per evaluation dimension.**

Exact and within-one-point agreement between the two non-clinician raters before adjudication, shown by scorecard dimension and overall.

| Non-clinician PLT Exact Inter-Rater Agreement (1-6 Scale) | |
|---|---|
| **Category** | **Agreement (%)** |
| Medical Jargon | 35.0% |
| Clarity | 57.5% |
| Fluency | 35.0% |
| Coherence | 52.5% |
| Relevance | 57.5% |
| User Trust | 30.0% |
| **Overall** | **44.6%** |

| Non-clinician PLTWithin-one-point Inter-Rater Agreement (±1, 1-6 Scale) | |
|---|---|
| **Category** | **Agreement (%)** |
| Medical Jargon | 75.0% |
| Clarity | 92.5% |
| Fluency | 67.5% |
| Coherence | 77.5% |
| Relevance | 90.0% |
| User Trust | 65.0% |
| **Overall** | **77.9%** |

**A.3.2: Non-clinician inter-rater agreement before adjudication for rPLT per evaluation dimension.**

Exact and within-one-point agreement between the two non-clinician raters before adjudication, shown by scorecard dimension and overall.

| **Non-clinician rPLT Exact Inter-Rater Agreement (1-6 Scale)** | |
|---|---|
| **Category** | **Agreement (%)** |
| Medical Jargon | 30.0% |
| Clarity | 42.5% |
| Fluency | 52.5% |
| Coherence | 55.0% |
| Relevance | 57.5% |
| User Trust | 37.5% |
| **Overall** | **45.8%** |

| **Non-clinician rPLT Within-one-point Inter-Rater Agreement (±1, 1-6 Scale)** | |
|---|---|
| **Category** | **Agreement (%)** |
| Medical Jargon | 70.0% |
| Clarity | 97.5% |
| Fluency | 87.5% |
| Coherence | 80.0% |
| Relevance | 92.5% |
| User Trust | 72.5% |
| **Overall** | **83.3%** |

**A.4: Substantial clinician disagreements before adjudication.**

Pre-adjudication clinician rating disagreements by scorecard dimension, defined as an absolute difference of 3 or more points on the six-point scale across PLT and rPLT evaluations.

| Clinician PLT Inter-Rater Disagreement (≥3, 1-6 Scale) | | |
|---|---|---|
| **Category** | **n** | **%** |
| Diagnosis Accuracy | 1 | 2.5% |
| Question Sufficiency for Clinical Safety | 1 | 2.5% |
| Triage Advice / Safety Accuracy | 0 | 0.0% |
| Medical Jargon | 0 | 0.0% |
| Clarity | 0 | 0.0% |
| Fluency | 0 | 0.0% |
| Coherence | 0 | 0.0% |
| Relevance | 0 | 0.0% |
| User Trust | 0 | 0.0% |
| Completeness | 0 | 0.0% |
| SBAR Adherence | 0 | 0.0% |
| Prudence | 0 | 0.0% |
| Brevity | 0 | 0.0% |
| **Overall** | **2** | **0.4%** |

| Clinician rPLT Inter-Rater Disagreement (≥3, 1-6 Scale) | | |
|---|---|---|
| **Category** | **n** | **%** |
| Diagnosis Accuracy | 2 | 5.0% |
| Question Sufficiency for Clinical Safety | 0 | 0.0% |
| Triage Advice / Safety Accuracy | 3 | 7.5% |
| Medical Jargon | 0 | 0.0% |
| Clarity | 0 | 0.0% |
| Fluency | 0 | 0.0% |
| Coherence | 1 | 2.5% |
| Relevance | 0 | 0.0% |
| User Trust | 4 | 10.0% |
| Completeness | 3 | 7.5% |
| SBAR Adherence | 1 | 2.5% |
| Prudence | 5 | 12.5% |
| Brevity | 0 | 0.0% |
| **Overall** | **19** | **3.7%** |

**A.5: Substantial non-clinician disagreements before adjudication.**

Pre-adjudication non-clinician rating disagreements by scorecard dimension, defined as an absolute difference of 3 or more points on the six-point scale across PLT and rPLT evaluations.

| **Non-clinician PLT Inter-Rater Disagreement (≥3, 1-6 Scale)** | | |
|---|---|---|
| **Category** | **n** | **%** |
| Medical Jargon | 5 | 12.5% |
| Clarity | 0 | 0.0% |
| Fluency | 1 | 2.5% |
| Coherence | 2 | 5.0% |
| Relevance | 2 | 5.0% |
| User Trust | 6 | 15.0% |
| **Overall** | **16** | **6.7%** |

| **Non-clinician rPLT Inter-Rater Disagreement (≥3, 1-6 Scale)** | | |
|---|---|---|
| **Category** | **n** | **%** |
| Medical Jargon | 0 | 0.0% |
| Clarity | 0 | 0.0% |
| Fluency | 1 | 2.5% |
| Coherence | 2 | 5.0% |
| Relevance | 0 | 0.0% |
| User Trust | 5 | 12.5% |
| **Overall** | **8** | **3.3%** |

### A.6: ICPC-2 Chief complaint mapping results

To assess whether standardized coding of chief complaints affected triage performance, we performed an ICPC-2 sensitivity analysis using the 10,030-case ESI dataset. Chief complaints were mapped to ICPC-2 codes using a codebook-constrained, LLM-assisted approach. When multiple complaints were present, they were considered jointly and assigned a single ICPC-2 code whenever possible; otherwise, the most clinically important complaint determined the primary code. In total, 10,015 cases were successfully assigned an ICPC-2 code. All other model inputs, model settings, data splits, and evaluation procedures were kept unchanged.

For rPLT, replacing the original chief complaint representation with ICPC-2 codes produced similar overall performance. Exact accuracy remained essentially unchanged, decreasing from 42.841% to 42.792%. Acceptable accuracy increased from 67.198% to 69.172%, undertriage decreased from 29.771% to 28.385%, and overtriage increased from 27.388% to 28.824%. These findings indicate a modest shift toward more conservative triage without a material change in exact performance.

For PLT, replacing the raw chief complaint text with the corresponding ICPC-2 code and rubric produced larger changes. Exact accuracy increased from 57.468% to 60.060%, whereas acceptable accuracy decreased from 86.122% to 71.924%. Undertriage increased from 13.071% to 27.637%, while overtriage decreased from 29.462% to 12.303%. One possible explanation is that the rule-based structure of rPLT constrains the influence of the chief complaint representation, whereas PLT relies more directly on the raw complaint text for clinical reasoning.

Overall, standardized ICPC-2 coding did not materially alter rPLT performance but substantially affected PLT. Because ICPC-2 was developed primarily for primary care and represents ED chief complaints at a broader level of abstraction, it may reduce the clinical granularity of heterogeneous ED presentations. Clinically distinct complaints may therefore map to the same broad ICPC-2 category. We consequently retained the original chief complaint representations in the primary analyses.

### A.7: Downstream clinical outcome triage validation results

To examine whether downstream clinical outcomes could provide additional context for evaluating triage appropriateness, we performed an exploratory analysis using linked MIMIC-IV ED, hospital, and ICU data. However, downstream outcomes do not necessarily indicate whether the initial triage decision was correct. Initial ESI assignment is based on the patient's presentation, apparent acuity, and anticipated resource needs at the time of arrival, whereas the eventual outcome may differ substantially after diagnostic evaluation and treatment. For example, a patient presenting with stroke-like symptoms may appropriately receive high-acuity triage even if subsequent evaluation excludes stroke and the patient is ultimately discharged home.

Downstream outcomes are also influenced by treatment response, disease progression, and local hospital resources and admission practices. For example, a patient who requires ICU admission at one institution may be managed outside the ICU at another institution with greater ward-based monitoring or specialist capacity. Outcome categories such as hospital admission and ICU admission may therefore not be standardized across institutions.

Encounters were classified as death, hospital admission with a long ICU stay (≥8 days), hospital admission with a short ICU stay (≤7 days), hospital admission without ICU care, or discharge home. The threshold for short ICU stay was based on Lefering et al. (2024). Recorded MIMIC-IV ESI levels were compared with a prespecified outcome-to-ESI mapping: death and long ICU stay were considered concordant with ESI 1–2, short ICU stay with ESI 2–3, hospital admission without ICU care with ESI 3–4, and discharge home with ESI 4–5.

The linked analytic cohort was restricted to 10,000 encounters resulting in hospital admission. Consequently, no discharge-home cases were present. The discharge-home category was retained to show the complete prespecified outcome-to-ESI framework, but concordance, undertriage, and overtriage could not be assessed for this category.

Overall, apparent undertriage was uncommon (67/10,000, 0.67%), whereas apparent overtriage occurred in 4,792 encounters (47.92%). Apparent overtriage was primarily observed among patients admitted without ICU care who had recorded ESI levels of 1 or 2. Concordance was highest for the most severe downstream outcomes, including death (237/281, 84.34%) and long ICU stay (142/159, 89.31%).

Supplementary Table 1**.** Exploratory downstream clinical outcome analysis using recorded MIMIC ESI.

| **Downstream clinical outcome** | **Prespecified concordant ESI range** | **Total encounters, n (%)** | **ESI distribution, n** | | | | | **Concordant, n (%)** | **Undertriage, n (%)** | **Overtriage, n (%)** |
|---|---|---|---|---|---|---|---|---|---|---|
| | | | **ESI 1** | **ESI 2** | **ESI 3** | **ESI 4** | **ESI 5** | | | |
| Death | 1–2 | 281 (2.81%) | 121 | 116 | 44 | 0 | 0 | 237 (84.34%) | 44 (15.66%) | 0 (0.00%) |
| Hospital admission with long ICU stay (≥8 days) | 1–2 | 159 (1.59%) | 61 | 81 | 17 | 0 | 0 | 142 (89.31%) | 17 (10.69%) | 0 (0.00%) |
| Hospital admission with short ICU stay (≤7 days) | 2–3 | 1,543 (15.43%) | 421 | 875 | 245 | 2 | 0 | 1,120 (72.59%) | 2 (0.13%) | 421 (27.28%) |
| Hospital admission without ICU | 3–4 | 8,017 (80.17%) | 581 | 3,79 | 3,607 | 35 | 4 | 3,642 (45.43%) | 4 (0.05%) | 4,371 (54.52%) |
| Discharge home | 4–5 | 0 (0.00%) | 0 | 0 | 0 | 0 | 0 | Not assessable | Not assessable | Not assessable |
| **Overall** | – | **10,000 (100.00%)** | **1,184** | **4,862** | **3,913** | **37** | **4** | **5,141 (51.41%)** | **67 (0.67%)** | **4,792 (47.92%)** |

These findings provide additional context regarding the relationship between initial ESI assignment and subsequent clinical course. However, they should not be interpreted as a direct validation of initial triage appropriateness, because a correct triage decision may be followed by a mild or substantially different downstream outcome. In addition, hospitalization and ICU utilization are affected by subsequent clinical management and institutional practices. Because the analysis included only admitted encounters, discharge-home outcomes and most of the low-acuity ESI 4–5 spectrum could not be evaluated. The analysis was therefore considered exploratory.

**Supplementary Information B: Prompting and LLM-as-judge strategies**

PLT, its patient simulator, and the LLM-judge were implemented using GPT-5.2. The text-generating steps of rPLT and LLM-assisted feature extraction were implemented using GPT-4o, except that GPT-5.2 was used for the SBAR summaries rated in the human evaluation and for all rPLT LLM steps in the must-not-miss simulations. Models were accessed through an institutional Azure OpenAI deployment. Outputs were generated between March 1, 2026 and May 22, 2026, except for the must-not-miss simulations and the Claude Sonnet 4.6 judge sensitivity analysis, which were run in October 2026. Full prompt templates and implementation files are available in the accompanying GitHub repositories: https://github.com/AutoEvaluation/TriageAIDev and https://github.com/AutoEvaluation/TriageAIEval. TriageAIDev contains the prompted LLM triage (PLT), rule-augmented prompted LLM triage (rPLT), and patient simulation components. TriageAIEval contains the reference-free LLM-as-judge and reference-based LLM-as-judge evaluation components.

For the 40 clinical vignettes used in human evaluation, all generated triage interactions and outputs were screened before evaluation for gross inconsistencies or malfunction. This screening assessed whether each run produced an interpretable triage interaction, remained on task, and generated outputs in the expected format. No outputs showed major malfunction requiring exclusion, such as irrelevant responses, failure to complete the triage task, or substantially incomplete interactions.

Supplementary Table 2: Input and output structure by LLM component

| LLM component | Input provided to the model | Output generated |
|---|---|---|
| Prompted LLM triage (PLT) | Patient chief complaint and simulated patient responses performed by the LLM patient simulator during the triage interaction | Follow-up questions, diagnosis and differential diagnosis, patient-facing care recommendation, and clinician-facing SBAR summary with predicted ESI |
| Rule-augmented prompted LLM triage (rPLT) | For acuity estimation: structured case variables (age, sex, hypertension, diabetes, chief complaint indicators, and binary symptom indicators). For the triage interaction: patient chief complaint and yes/no responses to follow-up questions | Predicted ESI from the two-stage statistical model; LLM-generated wording of follow-up questions, patient-facing care recommendation, diagnosis and differential diagnosis, and clinician-facing SBAR summary |
| LLM patient simulator | Structured MIMIC-IV derived clinical vignette information for the corresponding case including sex, age, chief complaint, complaint summary derived from history of present illness, medical history, family history | Patient-style responses to PLT follow-up questions and yes/no responses to rPLT follow-up questions, restricted to information available in the case record |
| Reference-free LLM-as-judge | Triage conversation, original tool output including patient-facing triage recommendation, diagnosis and differential diagnosis and SBAR summary with | Scores on the same six-point scorecard dimensions used in human evaluation |
| | predicted ESI level, and the complete scorecard rubric | |
| Reference-based LLM-as-judge | Clinician-generated reference outputs derived from rPLT outputs, triage conversation, original tool output including patient-facing triage recommendation, diagnosis and differential diagnosis and SBAR summary with predicted ESI level, and the complete scorecard rubric | Scores on the same six-point scorecard dimensions, using the clinician-generated reference as an additional scoring reference |

Reference-free LLM-as-judge evaluation was performed for both PLT and rPLT. Reference-based LLM-as-judge evaluation was performed only for rPLT because clinician-generated reference outputs were created only for rPLT. For LLM-as-judge evaluation, each case and scorecard dimension was scored across 7 repeated runs, and the median score was retained as the final LLM-judge score.

Repeated LLM-judge evaluations showed high scoring stability, with a mean standard deviation of 0.35 across repeated runs. No evaluation dimension had a standard deviation greater than the predefined instability threshold of 0.5. The highest standard deviations were observed for completeness (SD = 0.49), coherence (SD = 0.45), and medical jargon (SD = 0.43), although all remained below the instability threshold.

**Supplementary Information C: NLP-based chief complaint normalization and mapping**

To support clinical vignette selection and downstream triage analyses, we developed an NLP-based pipeline to normalize free-text chief complaints from the MIMIC-IV-ED triage table into reproducible, clinically interpretable categories. This procedure was used for case filtering, stratification, cohort characterization, and manual review. It was not used to define the reference triage outcome; the ground-truth ESI acuity label was retained directly from the original MIMIC-IV-ED acuity field.

Raw chief complaint text was first standardized before modeling. Leading and trailing whitespace or punctuation was removed, all text was converted to lowercase, repeated whitespace was collapsed, and empty or missing values were flagged. Common emergency department abbreviations were expanded into clinically interpretable terms, including “sob” to “shortness of breath,” “cp” to “chest pain,” “n/v” to “nausea and vomiting,” “ha” to “headache,” “s/p” to “status post,” “uti” to “urinary tract infection,” “etoh” to “alcohol,” and “mvc” to “motor vehicle collision.” After normalization, identical chief complaint strings were deduplicated, and their encounter-level frequencies were calculated.

Each unique normalized chief complaint was then mapped to a structured clinical concept using embedding-based similarity. Normalized chief complaint strings and PheCode description strings were embedded using the pretrained clinical language model Bio_ClinicalBERT. For each input, final-layer token embeddings were mean-pooled using the attention mask to generate a single vector representation. Chief complaint vectors and PheCode-description vectors were L2-normalized, and cosine similarity was computed between each chief complaint and each PheCode description. For each chief complaint, the PheCode with the highest cosine similarity was retained as the primary mapped concept; the next two highest-scoring PheCodes were also stored to support audit and manual review.

The chief complaint-to-PheCode mapping was merged back into the encounter-level triage table. The resulting dataset included the original chief complaint, normalized chief complaint, matched PheCode, matched PheCode description, PheCode category, and cosine similarity score for each encounter. These structured descriptors enabled cases to be filtered and grouped using clinically meaningful categories rather than raw free-text chief complaint strings.

During manual review, we identified a recurrent embedding-mapping failure mode in which the back-pain category attracted anatomically related but clinically distinct complaints. To address this issue, we applied a transparent rule-based post-processing step among complaints initially mapped to the back-pain category. Complaints containing body-site or symptom-specific keywords were reassigned to more appropriate PheCode-based categories, including cervicalgia, sciatica, abdominal pain, joint pain, limb pain, chest pain, neurologic or musculoskeletal symptoms, and selected breast, ear, eye, dental, pelvic, headache, or generalized pain categories. Complaints that did not meet any reassignment rule remained in the original back-pain category.

The final output was a reproducible encounter-level mapping from free-text chief complaints to structured PheCode-based clinical categories. This mapping supported clinical case filtering, cohort characterization, manual review, and vignette selection, while preserving the original MIMIC-IV-ED ESI acuity labels as the reference standard for triage evaluation.

## Supplementary Information D: Human evaluation scorecards and scoring protocols

This Supplementary Information provides the full human evaluation scorecards used by clinician and non-clinician raters, including rating dimensions, score anchors, and scoring instructions. It also describes the associated scoring protocols, including calibration, independent rating, adjudication procedures, and the process used to generate clinician-generated reference outputs for reference-based evaluation.

## D.1: Human evaluation scorecard for clinician raters

| | Medical Outcome Quality | | | User Interaction and Dialogue Evaluation (this section evaluates conversational performance only, not medical context) | | | | | | EHR Summary Evaluation | | | |
|---|---|---|---|---|---|---|---|---|---|---|---|---|---|
| | Diagnosis Accuracy | Question sufficiency for clinical safety | Triage Advice Safety / Accuracy | Medical Jargon | Clarity | Fluency | Coherence | Relevance | User Trust | Completeness | SBAR Adherence | Prudence | Brevity |
| Level | How accurately and thoroughly the AI identifies the most likely condition(s) based on the presented symptoms and information. | Measures whether the AI asked enough and the right questions to support a safe, defensible working diagnosis and avoid missing key conditions. | How appropriate the AI's recommended level of care or next action is given the clinical scenario (e.g., self-care, urgent care, ER, call ambulance). | Appropriateness of medical terminology for the user's likely health literacy level, including whether terms are explained when necessary. | How easily the note can be understood; absence of ambiguity | Smoothness of text flow within dialogue conversation | Logical linguistic connection and linguistic consistency within and between responses | How well the questions and elaborations in the chat are directly related to the conversation (content-wise and not linguistic) | How likely are you going to trust the AI's output based on the interaction from the user's (patient) perspective? | How complete and clinically coherent the summary is, based on the dialogue conversation | The LLM's ability to structure the summary evaluation in classic SBAR (Situation, Background, Assessment, Recommendation) format | The LLM's ability to express uncertainty or withhold conclusions when faced with incomplete, ambiguous, or out of scope information | Concise without losing meaning |
| 6 | Primary diagnosis is highly accurate; appropriate differentials included; no clinically important conditions missed. | All key questions were asked. No core details, red flags or key risks were omitted | Recommended level of care is fully appropriate for symptom severity; clear, safe, and actionable guidance. | Terminology perfectly calibrated to user level; medical terms clearly explained when used. | Completely clear and easy to understand; no ambiguity. | Conversation flows naturally and smoothly; fully human-like. | Responses are logically connected and fully consistent across the dialogue. No linguistic contradictions | All questions and responses are highly relevant and targeted to the user's symptoms and context. | Output is clear, confident, and clinically reasonable. I would fully trust the response and follow the triage recommendation. | All relevant symptoms, context, and timeline from the conversation are included. Information is clinically consistent with triage output. | Clearly structured in SBAR format. Situation, background, assessment, and triage recommendation are all present and well organized. | The AI is consistently appropriately cautious. It recognizes uncertainty, and avoids unsupported claims. | Concise and efficient with no unnecessary content, and no meaningful omissions. |

| | | | | | | | | | | | | | | |
|---|---|---|---|---|---|---|---|---|---|---|---|---|---|---|
| 5 | Diagnosis is largely accurate or clinically equivalent, not perfect but the next steps in diagnosis and treatment are almost the same | Almost all key key questions asked; minor non-critical gaps only | Triage is appropriate with only minor conservatism or mild over/under-triage; unlikely to cause harm. | Appropriate terminology with minor overuse or minor missing explanations. Does not meaningfully reduce understanding | Generally clear with only a minor ambiguous phrase or wording issue. | Mostly smooth with minor awkward phrasing or stiffness. | Minor logical gaps but overall consistent, or small continuity issues. | Mostly relevant with minor unnecessary content. | Response is mostly clear and appropriate with only minor concerns. I would likely follow the triage recommendation. | Nearly all relevant clinical details included with only minor omissions. Still supports the conversation's triage output. | All SBAR elements are present but organization or clarity could be slightly improved, but clinical usability remains very high | The AI is prudent overall, only minor overconfidence or slight unnecessary caution that does not meaningfully affect safety or appropriateness. | Slightly verbose but well controlled, without reducing usability |
| 4 | Diagnosis is reasonable but somewhat generic or missing relevant differentials; low risk of clinical impact. | Questions asked are mostly ok. Only one key area missed (core detail, red flag or key risk). Overall assessment remains mostly usable | Generally reasonable but somewhat miscalibrated; minor risk of inappropriate care level. | Some unnecessary jargon or inconsistent explanations but still understandable. | Understandable but several mildly vague or imprecise parts. | Understandable but somewhat mechanical, uneven or slightly unnatural in places. | Some inconsistencies or mild jumps in logic, or weak transitions. | Generally on topic but includes some tangential questions. | Response seems generally reasonable but has some uncertainty or missing clarity. I might follow the triage advice but would want additional confirmation. | Most key information from the conversation is present but one or more important details missing. Overall clinical picture is mostly clear. | Most SBAR components included, but one section is underdeveloped or not clearly separated. | The AI is generally reasonable, but there are some moments of mild overconfidence, under-caution, or insufficient acknowledgment of uncertainty. | Noticeably wordy or somewhat inefficient but key information remains easy to identify and the overall clinical usefulness is preserved.. |
| 3 | Partially correct or in the right direction but notable omissions; some risk of misinterpretation. | Questions asked may lead to a reasonable conclusion, but miss multiple significant details or risks | Triage level is questionable, noticeable mis-triage risk (e.g., too conservative or insufficiently urgent). | Frequent use of unexplained medical terminology., making parts of the conversation hard to understand | Meaning can still be followed, but frequent unclear phrasing requiring effort to interpret. | Noticeable awkwardness that affects readability and conversational flow. | Noticeable logical breaks or context misses, or inconsistencies that reduce quality of interaction. | Multiple irrelevant or poorly targeted questions. | Response raises noticeable concerns or lacks clarity. I would hesitate to follow the triage recommendation without outside guidance. | Core symptoms captured but several important details missing or slightly inconsistent. Triage direction based on the conversation may be uncertain in the summary . | Some SBAR elements present but structure is weak or multiple sections are incomplete. Usability is clearly reduced | The AI shows noticeable problems with calibration. It is too certain despite limited information, or is overly cautious in a way that reduces usefulness. | Excessively verbose or overly brief in places. |
| 2 | Diagnosis directionally related but ultimately off-target; important conditions or key features are overlooked, creating moderate risk of misinterpretation. | Big gaps in information gathered by AI. No core questions asked or no red flag explored or no risks screened., making safe triage unreliable | Triage recommendation is inappropriate and could delay proper care or lead to exessive amounts of care. | Heavy or inappropriate jargon that likely confuses many users. | Often confusing or ambiguous, and important parts are difficult to interpret confidently | Frequent disfluencies or unnatural phrasing that disrupt understanding. | Frequent inconsistencies or poor continuity, that confuse the interaction. | Significant portions of the dialogue are off-topic, Potential follow-up questions are missing | Response appears unreliable or confusing. I would not follow the triage advice and would seek other medical guidance. | Significant clinical information missing or inconsistent; summary output risks incorrect interpretation and delay of proper care | SBAR structure largely missing; important components such as assessment or recommendation are unclear. | The AI is poorly calibrated and gives conclusions with clearly inappropriate certainty or caution, creating meaningful clinical concern. | Length significantly harms usability. |
| 1 | Diagnosis clearly incorrect, misleading, or potentially harmful. | Minimal or irrelevant questions asked. Can not support triage or Dx | Dangerously incorrect triage (e.g., reassures when emergency care is needed) | Terminology is inaccessible, misleading, or inappropriate for patient communication. | Very difficult to understand overall. | Very difficult to follow; broken or unnatural language. | Dialogue is disjointed or internally contradictory to the point that it fails as a coherent interaction. | Dialogue largely fails to address the user's concern. Asks questions that have already been answered or does not follow-up on positive symptoms | Response is clearly unsafe or misleading. I think that following the triage recommendation could cause harm. | Critically incomplete or misleading summary likely to lead to harmful triage advice. | Does not follow SBAR structure and lacks clear triage framing. | The AI is recklessly overconfident, dangerously dismissive, or otherwise clearly imprudent in a way that could contribute to harm. | Extremely verbose or critically lacking essential detail. |

## D.2: Human evaluation scorecard for non-clinician raters

| Level | User Interaction and Dialogue Evaluation (this section evaluates conversational performance only, not medical context) | | | | | |
|---|---|---|---|---|---|---|
| | **Medical Jargon** | **Clarity** | **Fluency** | **Coherence** | **Relevance** | **User Trust** |
| | Appropriateness of medical terminology for the user's likely health literacy level, including whether terms are explained when necessary. | How easily the note can be understood; absence of ambiguity | Smoothness of text flow within dialogue conversation | Logical linguistic connection and linguistic consistency within and between responses | How well the questions and elaborations in the chat are directly related to the conversation (content-wise and not linguistic) | How likely are you going to trust the AI's output based on the interaction from the user's (patient) perspective? |
| 6 | Terminology perfectly calibrated to user level; medical terms clearly explained when used. | Completely clear and easy to understand; no ambiguity. | Conversation flows naturally and smoothly; fully human-like. | Responses are logically connected and fully consistent across the dialogue. No linguistic contradictions | All questions and responses are highly relevant and targeted to the user's symptoms and context. | Output is clear, confident, and clinically reasonable. I would fully trust the response and follow the triage recommendation. |
| 5 | Appropriate terminology with minor overuse or minor missing explanations. Does not meaningfully reduce understanding | Generally clear with only a minor ambiguous phrase or wording issue. | Mostly smooth with minor awkward phrasing or stiffness. | Minor logical gaps but overall consistent, or small continuity issues. | Mostly relevant with minor unnecessary content. | Response is mostly clear and appropriate with only minor concerns. I would likely follow the triage recommendation. |
| 4 | Some unnecessary jargon or inconsistent explanations but still understandable. | Understandable but several mildly vague or imprecise parts. | Understandable but somewhat mechanical, uneven or slightly unnatural in places. | Some inconsistencies or mild jumps in logic, or weak transitions. | Generally on topic but includes some tangential questions. | Response seems generally reasonable but has some uncertainty or missing clarity. I might follow the triage advice but would want additional confirmation. |
| 3 | Frequent use of unexplained medical terminology., making parts of the conversation hard to understand | Meaning can still be followed, but frequent unclear phrasing requiring effort to interpret. | Noticeable awkwardness that affects readability and conversational flow. | Noticeable logical breaks or context misses, or inconsistencies that reduce quality of interaction. | Multiple irrelevant or poorly targeted questions. | Response raises noticeable concerns or lacks clarity. I would hesitate to follow the triage recommendation without outside guidance. |
| 2 | Heavy or inappropriate jargon that likely confuses many users. | Often confusing or ambiguous, and important parts are difficult to interpret confidently | Frequent disfluencies or unnatural phrasing that disrupt understanding. | Frequent inconsistencies or poor continuity, that confuse the interaction. | Significant portions of the dialogue are off-topic, Potential follow-up questions are missing | Response appears unreliable or confusing. I would not follow the triage advice and would seek other medical guidance. |
| 1 | Terminology is inaccessible, misleading, or inappropriate for patient communication. | Very difficult to understand overall. | Very difficult to follow; broken or unnatural language. | Dialogue is disjointed or internally contradictory to the point that it fails as a coherent interaction. | Dialogue largely fails to address the user's concern. Asks questions that have already been answered or does not follow-up on positive symptoms | Response is clearly unsafe or misleading. I think that following the triage recommendation could cause harm. |

**D.3: Scorecard scoring protocol with file 1 being the triage AI output and file 2 being the full patient-file**

Protocol 1: Scorecard Completion

Below are the instructions for evaluating the AI's performance using only the information provided in the triage summary and AI chat flow of each case:

1. Using only the information provided in AI output, review the case and evaluate the AI's performance. For each metric in the scorecard, write the score down in the 'score' section, that best reflects the AI's performance. You may add notes within each section wherever necessary, especially in cases of ambiguity, uncertainty, or notable behavior.
2. Add general comments as needed. There is also an Overall Notes section for feedback that does not fit into the individual metric categories (e.g., unusual observations, patterns, or general impressions).
3. You may look at the full patient case for your own interest, but do not use information from this or assume missing information in your scorecard metric evaluation. Focus only on evaluating what the AI did in the triage conversation output.
4. After reading the full patient case you can use information to add to the Overall Note section if you want, not required.
5. After finishing rating one case, perform protocol 2 (Gold Standard creation) for that case.
6. Repeat this process for all cases.

Do NOT look at patient case before evaluating the AI output

**D.4: Clinician gold standard reference creation protocol**

Protocol 2: Gold Standard Creation (Clinician Ground Truth)

Below are the instructions to create a clinician-generated "gold standard" version of the triage interaction to compare against AI outputs for automated evaluation.

General Principe: The Gold standard should represent what you (the clinician) would have asked or done instead of the AI, given the same starting point.

1. Review the conversation from start to finish. As you review the interaction, revise any questions that are unclear, clinically inappropriate, or poorly phrased, while keeping their original intent intact. Or add details that are missing from the summary.
2. Add questions where important clinical information is missing or where additional clarification is necessary for proper triage, and insert these questions at the most appropriate point in the conversation flow.

3. Remove questions that are redundant, irrelevant, or clinically unnecessary, and do not replace them unless a better or more appropriate version is needed.
4. Do not edit questions that are already clear and clinically appropriate, and avoid making **minor** stylistic changes that do not affect meaning. Focus on clinical accuracy and completeness rather than perfect wording. (see example below)
5. Review the full interaction again to ensure it is complete, logically ordered, and clinically sound, without missing or redundant steps.
6. For the SBAR summary, add details that you find missing from the summary, or remove parts that you find non factual or cluttering the overall structure of the summary.
7. Repeat this process for all cases.

You may directly make edits to the triage conversation and summary in the output file. Remove any that you have changed and are not useful anymore to the case. In the end, we will analyze the parts of the text that have changed using software.

Protocol Step 2.4 Example:

Things to edit and things to not edit:

Do not edit:

"Are you having trouble breathing right now?" → "Do you find it difficult to breathe?"

Edit:

"Does your chest feel weird?" → Elaborate to "Are you experiencing chest pain, pressure, or tightness?"

**Supplementary Information E: Distributions of continuous vignette characteristics**

The age distribution across the 40 clinical vignettes was examined graphically to select summary statistics for Table 3. Age showed two modes, and the Shapiro–Wilk test indicated departure from normality (W = 0.93, P = 0.021). Age is therefore summarized as the median and interquartile range.

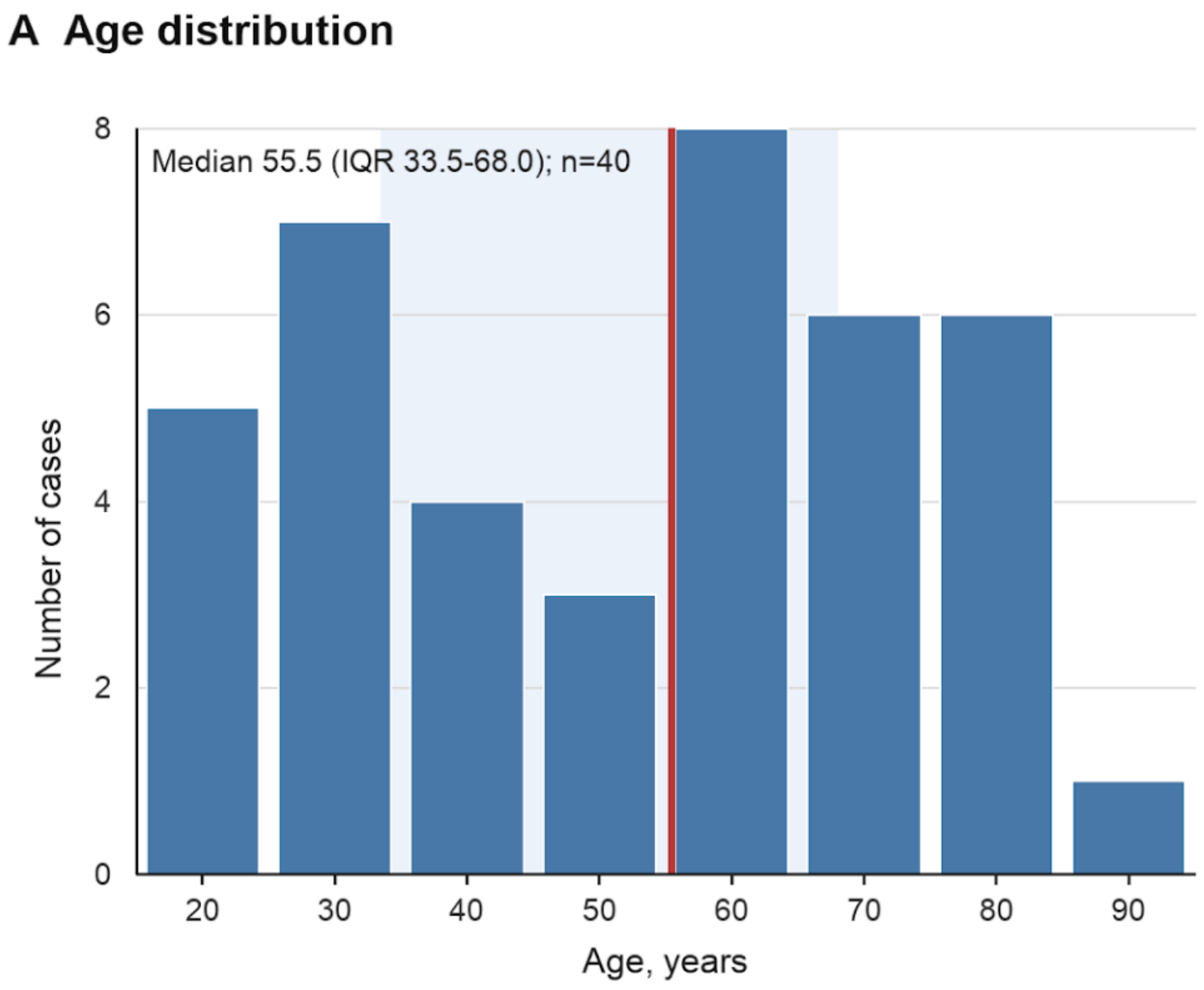

Supplementary Figure 1. Supplementary Figure 1. Distribution of age across the 40 clinical vignettes. Age is shown in years using 10-year bins centered on the axis labels. The red line indicates the median, and the shaded band indicates the interquartile range (IQR).

**Supplementary Information F: ESI classification by presenting complaint**

The 10,030 encounters in the triage classification analysis were sampled without restriction on chief complaint. To examine performance outside the nine complaint categories represented in the 40 clinical vignettes, encounters were grouped post hoc by keyword matching on the free-text chief complaint, including abbreviations common in triage documentation. Trauma-related complaints were identified by terms for falls, trauma, injury, motor vehicle or motorcycle collision, pedestrian struck, laceration, fracture, assault, head strike, stab or gunshot wound, and burn. Acute neurological complaints were identified by terms for seizure, stroke or transient ischemic attack, facial droop, slurred speech, dysarthria, aphasia or other speech disturbance, numbness, unilateral or focal weakness, hemiparesis, visual disturbance, diplopia, gait disturbance, and ataxia. The vignette complaint categories were abdominal pain, chest pain, dyspnea, fever, headache, nausea or vomiting, back pain, dizziness or lightheadedness, and altered mental status or confusion; all remaining encounters were classified as other complaints. The trauma-related and acute neurological groups were not mutually exclusive with the other groups, and nine encounters met both definitions.

PLT performance for trauma-related complaints was similar to its overall performance, whereas rPLT undertriage was higher for trauma-related complaints and for complaints outside the vignette categories. Undertriage was higher with both tools for acute neurological complaints, most of which had a reference ESI level of 1 or 2. These analyses were exploratory, and group assignment was not clinically adjudicated.

Supplementary Table 3. ESI classification performance of PLT and rPLT by presenting complaint group in the 10,030-case analysis. Values are percentages unless stated otherwise. Acceptable accuracy was defined as exact ESI agreement or prediction of one level more conservative than the reference label.

| **Complaint group** | **n** | **Reference ESI 1–2** | **Tool** | **Exact** | **Acceptable** | **Undertriage** | **Overtriage** |
|---|---|---|---|---|---|---|---|
| All encounters | 10,030 | 60.4 | PLT | 57.5 | 86.1 | 13.1 | 29.5 |
| | | | rPLT | 42.8 | 67.2 | 29.8 | 27.4 |
| Vignette complaint categories | 5,097 | 59.7 | PLT | 57.7 | 88.0 | 11.4 | 30.9 |
| | | | rPLT | 43.5 | 76.7 | 19.8 | 36.8 |
| Other complaints | 4,933 | 61.1 | PLT | 57.2 | 84.2 | 14.8 | 28.0 |
| | | | rPLT | 42.2 | 57.4 | 40.1 | 17.7 |
| Trauma-related complaints | 627 | 65.7 | PLT | 58.9 | 85.3 | 14.2 | 27.0 |
| | | | rPLT | 32.5 | 48.3 | 48.6 | 18.8 |
| Acute neurological complaints | 551 | 84.6 | PLT | 56.6 | 72.8 | 26.3 | 17.1 |
| | | | rPLT | 27.4 | 38.8 | 58.3 | 14.3 |

**Supplementary Information G: LLM-judge sensitivity analysis with a non-GPT model**

To assess dependence on the judge model, the LLM-judge evaluation was repeated with Claude Sonnet 4.6 (model identifier claude-sonnet-4-6) using the same rubric as the GPT-5.2 judge. Each of the 40 cases was evaluated seven times in each of three judging modes (PLT reference-free, rPLT reference-free, and rPLT reference-based), giving 840 evaluations, and the median across runs was retained for each dimension. The maximum output length was 4,096 tokens and the sampling temperature was left at the service default. In reference-based mode, the Claude judge received both clinician-generated reference outputs together, whereas GPT-5.2 scores were obtained against each reference separately and averaged. User trust was not scored with the Claude judge, so 12 dimensions were compared.
Mean scores across the 12 dimensions with the Claude judge were 4.51 for PLT reference-free, 3.71 for rPLT reference-free, and 4.00 for rPLT reference-based judging (Supplementary Table 4). In reference-free mode, PLT was scored higher than rPLT in 38 of 40 cases, lower in one, and equal in one (two-sided exact sign test, $p < 0.001$). The direction of the difference between tools agreed between the two judges in 11 of 12 dimensions (Supplementary Table 5). Across repeated runs, the mean standard deviation of Claude scores was 0.13, and 9 of 1,440 case–dimension combinations exceeded the instability threshold of 0.5. Correlations with clinician and non-clinician ratings are reported as signed Spearman coefficients across the 40 cases (Supplementary Tables 6 and 7); mean absolute coefficients for clinician ratings were 0.193, 0.230, and 0.254 with the Claude judge and 0.190, 0.186, and 0.223 with the GPT-5.2 judge for the three modes, and for non-clinician ratings 0.147, 0.082, and 0.125 with Claude and 0.087, 0.120, and 0.176 with GPT-5.2.

Supplementary Table 4. Mean LLM-judge scores by dimension for the Claude Sonnet 4.6 and GPT-5.2 judges. Scores are means across 40 cases on the 1 to 6 scale.

| **Dimension** | **PLT reference-free, Claude** | **PLT reference-free, GPT-5.2** | **rPLT reference-free, Claude** | **rPLT reference-free, GPT-5.2** | **rPLT reference-based, Claude** | **rPLT reference-based, GPT-5.2** |
|---|---|---|---|---|---|---|
| Diagnosis Accuracy | 4.400 | 3.975 | 3.800 | 2.875 | 4.225 | 3.788 |
| Question Sufficiency | 4.175 | 4.075 | 3.075 | 2.450 | 2.950 | 2.850 |
| Triage recommendation appropriateness | 5.025 | 4.875 | 4.425 | 4.150 | 4.425 | 4.138 |
| Medical Jargon | 4.950 | 3.650 | 5.800 | 5.375 | 5.775 | 5.287 |
| Clarity | 4.950 | 4.375 | 4.800 | 4.625 | 4.800 | 4.775 |
| Fluency | 4.925 | 3.950 | 4.125 | 3.450 | 4.150 | 3.313 |
| Coherence | 4.600 | 3.600 | 2.475 | 3.475 | 2.400 | 3.500 |
| Relevance | 5.125 | 4.125 | 2.825 | 3.375 | 2.800 | 3.462 |
| Completeness | 3.325 | 3.600 | 2.325 | 2.475 | 3.575 | 3.138 |
| SBAR Adherence | 4.000 | 4.675 | 3.050 | 3.925 | 4.525 | 4.638 |
| Prudence | 3.750 | 4.700 | 3.525 | 3.475 | 3.400 | 4.250 |
| Brevity | 4.875 | 4.175 | 4.275 | 4.075 | 4.950 | 4.150 |

Supplementary Table 5. Difference between PLT and rPLT in reference-free LLM-judge scores. Differences are PLT minus rPLT. For the Claude judge, the number of cases in which each tool was scored higher and the two-sided exact sign test are shown; for the GPT-5.2 judge, the difference in mean scores is shown.

| **Dimension** | **PLT, Claude** | **rPLT, Claude** | **Difference, Claude** | **Cases PLT higher** | **Cases rPLT higher** | **p value** | **Difference, GPT-5.2** |
|---|---|---|---|---|---|---|---|
| Diagnosis Accuracy | 4.400 | 3.800 | +0.60 | 19 | 6 | 0.015 | +1.10 |
| Question Sufficiency | 4.175 | 3.075 | +1.10 | 30 | 2 | <0.001 | +1.62 |
| Triage recommendation appropriateness | 5.025 | 4.425 | +0.60 | 19 | 4 | 0.003 | +0.72 |
| Medical Jargon | 4.950 | 5.800 | -0.85 | 0 | 32 | <0.001 | -1.73 |
| Clarity | 4.950 | 4.800 | +0.15 | 7 | 1 | 0.070 | -0.25 |
| Fluency | 4.925 | 4.125 | +0.80 | 32 | 0 | <0.001 | +0.50 |
| Coherence | 4.600 | 2.475 | +2.12 | 40 | 0 | <0.001 | +0.12 |
| Relevance | 5.125 | 2.825 | +2.30 | 39 | 0 | <0.001 | +0.75 |
| Completeness | 3.325 | 2.325 | +1.00 | 31 | 2 | <0.001 | +1.12 |
| SBAR Adherence | 4.000 | 3.050 | +0.95 | 30 | 2 | <0.001 | +0.75 |
| Prudence | 3.750 | 3.525 | +0.23 | 12 | 5 | 0.143 | +1.23 |
| Brevity | 4.875 | 4.275 | +0.60 | 25 | 1 | <0.001 | +0.10 |
| Mean of 12 dimensions | 4.508 | 3.708 | +0.80 | 38 | 1 | <0.001 | +0.50 |

Supplementary Table 6. Spearman correlations between LLM-judge scores and clinician ratings across 40 cases.

| **Dimension** | **PLT reference-free, Claude** | **PLT reference-free, GPT-5.2** | **rPLT reference-free, Claude** | **rPLT reference-free, GPT-5.2** | **rPLT reference-based, Claude** | **rPLT reference-based, GPT-5.2** |
|---|---|---|---|---|---|---|
| Diagnosis Accuracy | 0.616 | 0.538 | 0.167 | 0.479 | 0.318 | 0.394 |
| Question Sufficiency | 0.571 | 0.490 | 0.408 | 0.440 | 0.474 | 0.625 |
| Triage recommendation appropriateness | 0.081 | 0.168 | 0.376 | 0.312 | 0.496 | 0.730 |
| Medical Jargon | -0.104 | 0.099 | 0.198 | -0.137 | 0.189 | -0.083 |
| Clarity | -0.053 | -0.059 | 0.071 | 0.121 | 0.071 | 0.165 |
| Fluency | -0.164 | 0.132 | 0.011 | 0.010 | 0.312 | 0.052 |
| Coherence | 0.041 | -0.131 | 0.337 | -0.155 | 0.237 | 0.164 |
| Relevance | -0.007 | -0.025 | 0.566 | 0.030 | 0.458 | 0.126 |
| Completeness | -0.180 | -0.318 | 0.294 | 0.303 | 0.079 | 0.155 |
| SBAR Adherence | -0.114 | 0.061 | 0.166 | -0.078 | -0.005 | 0.057 |
| Prudence | 0.356 | 0.043 | 0.124 | 0.123 | -0.153 | 0.082 |
| Brevity | 0.025 | 0.212 | -0.039 | -0.043 | 0.254 | -0.042 |

Supplementary Table 7. Spearman correlations between LLM-judge scores and non-clinician ratings across 40 cases.

| **Dimension** | **PLT reference-free, Claude** | **PLT reference-free, GPT-5.2** | **rPLT reference-free, Claude** | **rPLT reference-free, GPT-5.2** | **rPLT reference-based, Claude** | **rPLT reference-based, GPT-5.2** |
|---|---|---|---|---|---|---|
| Medical Jargon | 0.060 | 0.128 | 0.095 | -0.128 | 0.061 | 0.041 |
| Clarity | -0.200 | 0.020 | 0.105 | 0.277 | 0.105 | -0.204 |
| Fluency | -0.009 | -0.207 | 0.128 | 0.045 | 0.212 | 0.206 |
| Coherence | -0.289 | 0.024 | -0.051 | 0.023 | -0.066 | 0.179 |
| Relevance | -0.175 | 0.056 | 0.030 | 0.126 | 0.181 | 0.249 |

**Supplementary Information H: Supplementary review by emergency physicians**

To assess the influence of rater specialty, two emergency physicians reviewed all 40 cases for both tools. They had access to the scores assigned by the critical care clinicians and assigned their own scores on the same 13 dimensions (1,040 ratings). Emergency physician scores were averaged without adjudication. Because the original scores were visible, this review was not an independent blinded validation.

The emergency physicians scored PLT higher than rPLT on all 13 dimensions and in 38 of 40 cases (two-sided exact sign test on case-level mean scores, $p < 0.001$; Supplementary Table 8). Their mean scores were lower than the critical care clinicians' scores by 0.34 points for PLT and 0.10 points for rPLT, with the largest differences for prudence and user trust. The mean emergency physician score was within one point of the critical care score for 95.0% of PLT ratings and 92.5% of rPLT ratings. Agreement between the two emergency physicians is shown in Supplementary Table 9; one emergency physician assigned systematically lower scores than the other (mean, 5.09 vs. 5.74 for PLT and 4.17 vs. 4.88 for rPLT).

Supplementary Table 8. Mean scores assigned by critical care clinicians and emergency physicians, by dimension. Scores are means across 40 cases on the 1 to 6 scale; critical care scores are the final adjudicated scores. The difference is PLT minus rPLT for emergency physician scores, with the number of cases in which each tool was scored higher.

| **Dimension** | **PLT, critical care** | **PLT, emergency** | **rPLT, critical care** | **rPLT, emergency** | **Difference, emergency** | **Cases PLT higher** | **Cases rPLT higher** |
|---|---|---|---|---|---|---|---|
| Diagnosis Accuracy | 5.29 | 5.09 | 4.44 | 3.76 | +1.32 | 35 | 3 |
| Question Sufficiency | 5.17 | 4.90 | 4.08 | 3.50 | +1.40 | 38 | 0 |
| Triage recommendation appropriateness | 5.35 | 4.99 | 4.25 | 4.06 | +0.93 | 33 | 5 |
| Medical Jargon | 5.83 | 5.47 | 5.00 | 5.29 | +0.19 | 15 | 2 |
| Clarity | 5.97 | 5.75 | 4.96 | 5.33 | +0.42 | 25 | 1 |
| Fluency | 5.88 | 5.60 | 4.60 | 5.00 | +0.60 | 33 | 3 |
| Coherence | 5.76 | 5.56 | 4.31 | 4.78 | +0.79 | 36 | 1 |
| Relevance | 5.91 | 5.50 | 4.28 | 4.36 | +1.14 | 36 | 0 |
| User Trust | 5.94 | 5.33 | 4.61 | 3.88 | +1.45 | 37 | 0 |
| Completeness | 5.92 | 5.54 | 4.75 | 4.96 | +0.57 | 22 | 3 |
| SBAR Adherence | 5.94 | 5.76 | 5.00 | 5.39 | +0.38 | 25 | 0 |
| Prudence | 5.91 | 5.12 | 4.85 | 4.11 | +1.01 | 34 | 3 |
| Brevity | 5.91 | 5.74 | 5.04 | 4.44 | +1.30 | 39 | 0 |
| Mean of 13 dimensions | 5.75 | 5.41 | 4.63 | 4.53 | +0.88 | 38 | 2 |

Supplementary Table 9. Agreement between the two emergency physicians, by dimension. Values are percentages of 40 cases with exact agreement and with agreement within one point on the 1 to 6 scale.

| **Dimension** | **PLT, exact** | **PLT, within one point** | **rPLT, exact** | **rPLT, within one point** |
|---|---|---|---|---|
| Diagnosis Accuracy | 20.0 | 85.0 | 25.0 | 82.5 |
| Question Sufficiency | 27.5 | 85.0 | 47.5 | 92.5 |
| Triage recommendation appropriateness | 27.5 | 75.0 | 35.0 | 67.5 |
| Medical Jargon | 10.0 | 90.0 | 2.5 | 70.0 |
| Clarity | 52.5 | 97.5 | 7.5 | 82.5 |
| Fluency | 45.0 | 92.5 | 0.0 | 50.0 |
| Coherence | 35.0 | 95.0 | 5.0 | 40.0 |
| Relevance | 35.0 | 95.0 | 20.0 | 70.0 |
| User Trust | 30.0 | 95.0 | 45.0 | 87.5 |
| Completeness | 45.0 | 97.5 | 15.0 | 82.5 |
| SBAR Adherence | 52.5 | 100.0 | 2.5 | 80.0 |
| Prudence | 7.5 | 65.0 | 30.0 | 70.0 |
| Brevity | 47.5 | 100.0 | 17.5 | 85.0 |
| All dimensions | 33.5 | 90.2 | 19.4 | 73.8 |

**Supplementary Information I: Undertriage and overtriage by reference ESI level and illustrative cases**

Supplementary Table 10 shows the distribution of exact agreement, undertriage, and overtriage by reference ESI level in the 10,030-case analysis. Undertriage was concentrated among encounters with a reference ESI level of 1. PLT undertriage in these encounters was almost entirely by one level, whereas rPLT assigned ESI level 3 or lower to 45.1% of them. Of all undertriaged encounters, 58.9% of those undertriaged by PLT were ESI level 1 encounters assigned ESI level 2, whereas 65.2% of those undertriaged by rPLT were ESI level 2 encounters assigned ESI level 3 and 18.3% were undertriaged by two or more levels.

Supplementary Table 10. ESI classification by reference ESI level in the 10,030-case analysis. Values are percentages of encounters at each reference level. Undertriage is a prediction less urgent than the reference level and overtriage a prediction more urgent than the reference level.

| **Reference ESI** | **n** | **PLT exact** | **PLT under-triage** | **PLT under-triage by ≥2 levels** | **PLT over-triage** | **rPLT exact** | **rPLT under-triage** | **rPLT under-triage by ≥2 levels** | **rPLT over-triage** |
|---|---|---|---|---|---|---|---|---|---|
| 1 | 1,185 | 32.9 | 67.1 | 1.9 | 0.0 | 14.9 | 85.1 | 45.1 | 0.0 |
| 2 | 4,872 | 82.8 | 10.4 | 0.0 | 6.8 | 44.1 | 40.2 | 0.2 | 15.7 |
| 3 | 3,930 | 34.1 | 0.2 | 0.0 | 65.6 | 50.2 | 0.5 | 0.0 | 49.3 |
| 4 | 39 | 0.0 | 0.0 | 0.0 | 100.0 | 0.0 | 0.0 | 0.0 | 100.0 |
| 5 | 4 | 0.0 | 0.0 | 0.0 | 100.0 | 0.0 | 0.0 | 0.0 | 100.0 |

Three cases from the 40 clinical vignettes illustrate undertriage and overtriage relative to recorded ESI levels. Each is described across 20 repeated runs of each tool.

A man in his sixties presented with worsening back pain; leg weakness and bowel incontinence were documented in the record (reference ESI level 1). With PLT, the simulated patient reported both symptoms in all 20 repeated runs, PLT listed a serious spinal cause in its output, and it assigned ESI level 2 in all 20 runs (one-level undertriage). With rPLT, questions about numbness, leg weakness, or bowel function were asked in all 20 runs, and no positive response was recorded; the generated primary diagnosis was low back pain, and the acuity model assigned ESI level 3 (two-level undertriage).

A woman in her twenties presented with fever, headache, and body aches; neck stiffness was documented in the record (reference ESI level 1). With PLT, neck stiffness was reported in 19 of 20 runs, meningitis was named in the output in all 20 runs, and ESI level 2 was assigned in all 20 runs (one-level undertriage). With rPLT, neck stiffness was asked about in 2 of 20 runs; meningitis was the primary diagnosis in 1 run and appeared in the differential diagnosis in 14, and the acuity model assigned ESI level 3 (two-level undertriage).

A woman in her forties presented with mental fogginess, dizziness, and chest discomfort after a recent medication change (reference ESI level 4). rPLT assigned ESI level 2 (two-level overtriage) and asked about medication in 1 of 20 runs. PLT assigned ESI level 3 in 17 runs and ESI level 2 in 3 runs (one-level or two-level overtriage), and the medication change was reported to PLT in 17 of 20 runs.

These examples reflect disagreement with recorded ESI levels and not independently adjudicated clinical errors. In rPLT, the acuity level is computed from structured case variables, so information elicited during the interaction does not change it.